\documentclass[10pt,twocolumn,letterpaper]{article}

\usepackage{iccv}
\usepackage{times}
\usepackage{epsfig}
\usepackage{graphicx}
\usepackage{amsmath}
\usepackage{amssymb}
\usepackage{booktabs}
\usepackage{multirow}
\usepackage{cuted}
\usepackage{color}
\definecolor{url_color}{RGB}{42, 83, 163}
\usepackage{caption}
\usepackage{eso-pic}
\usepackage{diagbox}
\usepackage[accsupp]{axessibility}

\usepackage[breaklinks=true,bookmarks=false]{hyperref}

\iccvfinalcopy 

\def\iccvPaperID{7107} 

\ificcvfinal\fi

\begin{document}

\title{IntrinsicNeRF: Learning Intrinsic Neural Radiance Fields for Editable Novel View Synthesis}

\author{
  Weicai Ye$^1$\thanks{indicates equal contribution.}
\and
  Shuo Chen$^1$\footnotemark[1]
  \and
  Chong Bao$^1$\and
  Hujun Bao$^1$\and
  Marc Pollefeys$^{2,3}$\and
  Zhaopeng Cui$^1$\and
  Guofeng Zhang$^1$\thanks{Corresponding author.}\and
  \textnormal{$^1$State Key Lab of CAD\&CG, Zhejiang University} \qquad \textnormal{$^2$ETH Zurich} \qquad
  \textnormal{$^3$Microsoft} 
  \\
  \textbf{\small \urlstyle{tt}\textcolor{url_color}{\url{https://zju3dv.github.io/intrinsic_nerf}}}
}

\twocolumn[{%
    \renewcommand\twocolumn[1][]{#1}%
    \setlength{\tabcolsep}{0.0mm} 
    \newcommand{\sz}{0.125}  
    \maketitle
    \begin{center}
        \newcommand{\teaserwidth}{\textwidth}
    \vspace{-3em}
        \includegraphics[width=\linewidth]{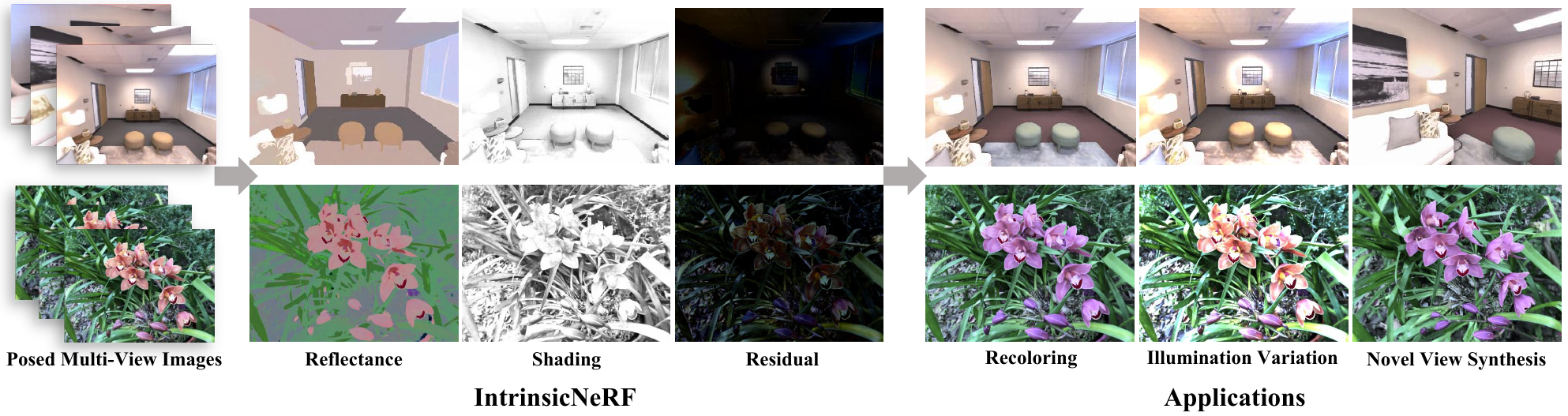}
      \vspace{-2em}
        \captionof{figure}{\textbf{Intrinsic Neural Radiance Fields (IntrinsicNeRF).} Given multi-view posed images of static scenes, IntrinsicNeRF can factorize them into the multi-view consistent components: reflectance, shading, and residual layers. The decomposition can support online applications such as scene recoloring, illumination variation, and editable novel view synthesis.}
    \label{fig:teaser}
    \end{center}
}]
\def\thefootnote{$\ast$}\footnotetext{ indicates equal contribution.}\def\thefootnote{$^\dagger$}\footnotetext{ indicates corresponding author.}

\ificcvfinal\thispagestyle{empty}\fi

\begin{abstract}
Existing inverse rendering combined with neural rendering methods can only perform editable novel view synthesis on object-specific scenes, while we present intrinsic neural radiance fields, dubbed IntrinsicNeRF, which introduce intrinsic decomposition into the NeRF-based neural rendering method and can extend its application to room-scale scenes. Since intrinsic decomposition is a fundamentally under-constrained inverse problem, we propose a novel distance-aware point sampling and adaptive reflectance iterative clustering optimization method, which enables IntrinsicNeRF with traditional intrinsic decomposition constraints to be trained in an unsupervised manner, resulting in multi-view consistent intrinsic decomposition results. To cope with the problem that different adjacent instances of similar reflectance in a scene are incorrectly clustered together, we further propose a hierarchical clustering method with coarse-to-fine optimization to obtain a fast hierarchical indexing representation. It supports compelling real-time augmented applications such as recoloring and illumination variation. Extensive experiments and editing samples on both object-specific/room-scale scenes and synthetic/real-word data demonstrate that we can obtain consistent intrinsic decomposition results and high-fidelity novel view synthesis even for challenging sequences. 
\end{abstract}

\vspace{-1.5em}
\section{Introduction}
\label{sec:intro}
Recently neural rendering techniques have gained increasing attention and demonstrated tremendous performance in novel view synthesis, ranging from small objects~\cite{liu2020neural, martin2021nerf, mildenhall2020nerf, wang2021neus} to large outdoor scenes~\cite{martin2021nerf, tancik2022block}, but they struggle to perform further intuitive editing like realistic scene recoloring, relighting, etc, for the scenes are usually represented as neural fields implicitly and required to be decomposed into the editable properties explicitly.

Several works have proposed to fulfill this goal by introducing inverse rendering into neural rendering~\cite{zhang2021physg,zhang2021nerfactor,zhang2022modeling}, where the scene is decomposed into geometry, reflectance, and illumination.
However, since inverse rendering is fundamentally ambiguous and highly ill-posed, these NeRF-based inverse rendering works~\cite{zhang2021physg, zhang2022modeling} introduce many prior assumptions preventing the modeling of mutual occlusion, inter-reflection, and indirect light propagation of different objects in the scene. An accurate 3D surface recovery is also required as a prerequisite. All these factors limit their applications to object-specific scenarios.

\begin{figure*}[t!]
    \centering
    \includegraphics[width=1.0\linewidth, trim={0 0 0 0}, clip]{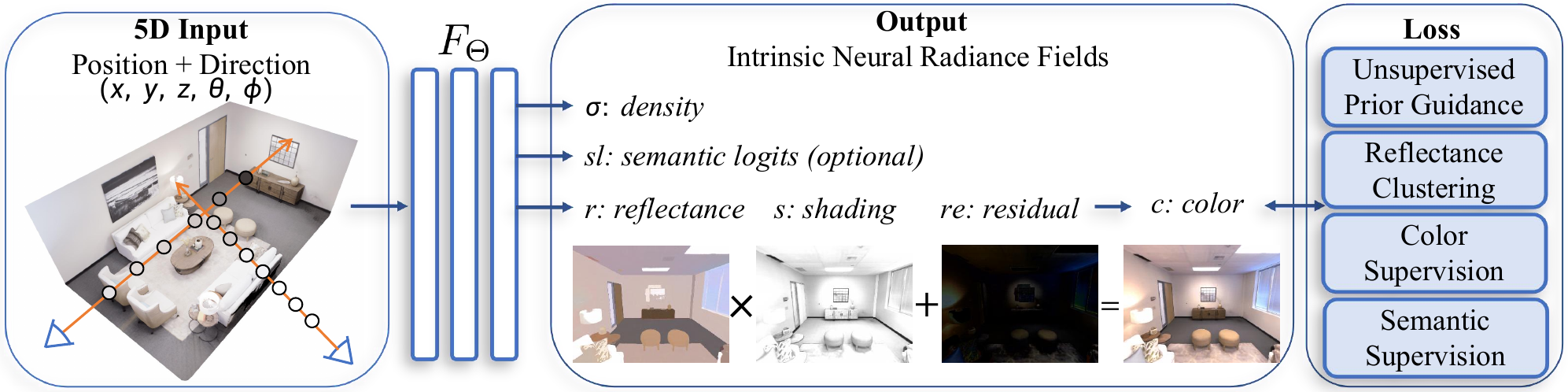}
    \vspace{-1.5em}
    \caption{\textbf{IntrinsicNeRF Framework.} IntrinsicNeRF takes the sampled spatial coordinate point and direction as input and outputs the density, reflectance, shading, and residual term. The semantic branch is optional. Unsupervised Prior and Reflectance Clustering are exploited as the loss function constraints to train the IntrinsicNeRF in an unsupervised manner. 
    }
    \label{fig: IntrinsicNeRF Framework}
\end{figure*}

To empower such editable capabilities to the scene-level neural rendering, we present intrinsic neural radiance fields, which introduce intrinsic decomposition into neural rendering, based on the fact that intrinsic decomposition can be considered as a simplification of inverse rendering designed to provide interpretable intermediate representations (i.e., reflectance and shading) that are relatively easy to solve for both in small objects and large scenes. A potential naive solution may use the trained NeRF model to generate multi-view images and then perform multi-view intrinsic decomposition, where these two tasks are separated. In contrast, extending from NeRF~\cite{mildenhall2020nerf}, IntrinsicNeRF (see Sec.~\ref{Sec: Intrinsic Neural Radiance Fields} and Fig.~\ref{fig: IntrinsicNeRF Framework}) takes the sampled spatial coordinate point $\mathbf{x} = (x,y,z)$ and the direction $\mathbf{d} = (\theta, \phi)$ as input and regresses them into the density $\sigma$, view-independent reflectance $r$ and shading $s$ (Lambertian reflectance assumption) and additional view-dependent residual term $re$~\cite{maxwell2008bi, tominaga1994dichromatic} (Eq.~\ref{equa: intrinsic residual}), which naturally guarantees the multi-view consistency of decomposition after training, thanks to neural rendering. 

However, it is nontrivial to design such a framework due to huge gaps in optimization between traditional intrinsic decomposition and NeRF-based methods. Traditional intrinsic decomposition methods optimize the energy equation by establishing constraints related to the image pixels, while NeRF-based methods optimize the view-dependent densities and colors of several sampled 3D points through volume rendering, which makes it hard to exploit the commonly used prior knowledge in intrinsic decomposition (see Sec.~\ref{Sec: Unsupervised Priors}) such as chromaticity prior, reflectance sparsity, etc.
To address this problem, we propose a distance-aware sampling method (see Fig.~\ref{fig: Distance-Aware Point Sampling}) that allows the sampled points not only to be random but also to establish local and global relationships between points. In this way, IntrinsicNeRF satisfies both the novel view synthesis and the better recovery of the intrinsic properties of the scene. 

Moreover, to deal with the inconsistencies of similar reflectance regions~\cite{meka2016live}, we present an adaptive reflectance iterative clustering method (see Sec.~\ref{Sec: Adaptive Reflectance Clustering with Mean Shift}) with mean shift~\cite{cheng1995mean} to adaptively cluster color points with similar reflectance based on the scene itself, rather than K-Means used in~\cite{meka2016live}, which limits the number of specific classes. A continuously updated clustering operation with the voxel grid filter is constructed to map similar reflectance colors to the same target reflectance color and then obtain the clustered category for each color point (see Fig.~\ref{fig: Reflectance Clustering with Mean Shift}). 
 
To settle the problem of different adjacent instances of similar reflectance in a scene being clustered together, we propose a semantic-aware reflectance sparsity constraint during training. Inspired by Semantic-NeRF~\cite{zhi2021place}, we add an additional semantic branch to IntrinsicNeRF, along with reflectance clustering, which yields a hierarchical reflectance iterative clustering and indexing method (see Fig.~\ref{fig: Hierarchical Clustering and Indexing}), optimizing the network from coarse to fine. Extensive experiments on the Blender Object and the Replica Scene dataset demonstrate our method can obtain consistent intrinsic decomposition results and high-fidelity novel view synthesis even for challenging sequences. 
We also develop video editing software to facilitate users to perform online scene recoloring, illumination variation, and editable novel view synthesis on both real-world and synthetic data on the CPU (see Fig.~\ref{fig:teaser}).

\section{Related Work}
\noindent\textbf{Intrinsic Image Decomposition.}\label{Sec:Intrinsic Image Decomposition} Intrinsic decomposition~\cite{barrow1978recovering} is a typical image layer separation problem aimed at decomposing images into reflectance, shading, etc., and has been studied for decades. To deal with this ill-posed problem, additional priors~\cite{horn1974determining, land1971lightness, shen2008intrinsic} with optimization framework have been used. 
Recently, deep learning methods~\cite{baslamisli2018joint, fan2018revisiting, liu2020learning, luo2020niid, yu2019inverserendernet, zhu2022irisformer} have emerged to perform intrinsic decomposition, and with large datasets~\cite{li2018cgintrinsics,li2018learning,roberts2021hypersim}, they have shown further improvement. Unsupervised intrinsic image decomposition works~\cite{janner2017self, liu2020unsupervised} have also achieved impressive results. 
IntrinsicNeRF considers not only the intrinsic decomposition prior but also the consistency of different perspectives in neural rendering, performing unsupervised optimization of the network.

\noindent\textbf{Intrinsic Video Decomposition.}
\label{Sec:Intrinsic Video Decomposition}
Intrinsic video decomposition extends intrinsic decomposition from the image domain to the video domain and can be roughly divided into two types. One is to perform the intrinsic image decomposition first and use the motion information to establish the correlation between frames for post-processing~\cite{bonneel2015blind, lai2018learning, ye2014intrinsic}. The other is to directly unify the image's local and global relations using some prior, by optimizing the energy equation~\cite{bonneel2014interactive, meka2016live}. There are also works~\cite{duchene2015multi, joy2022multi, laffont2012coherent, yi2020leveraging} on intrinsic decomposition from multi-view images. These methods have some consistency in intrinsic video decomposition but are unable to perform novel view synthesis. While IntrinsicNeRF introduces traditional intrinsic decomposition prior to the neural radiance fields to achieve end-to-end optimization, which not only performs better intrinsic video decomposition than previous methods but also allows for realistic editable novel view synthesis.

\noindent\textbf{Inverse Rendering.}
\label{Sec:Inverse Rendering} Inverse rendering~\cite{garces2022survey} is another way to restore the basic properties of scene, which can be broadly divided into two categories: classic approaches~\cite{blanz1999morphable, oh2001image, kanamori2019relighting}, differentiable renders~\cite{li2018differentiable, nimier2019mitsuba, zhao2020physics, loubet2019reparameterizing} methods.
Plenty of works combining neural rendering with inverse rendering~\cite{neural_outdoor_rerender, boss2021nerd, rudnev2022nerfosr, yang2022_nr_in_a_room, zhang2021physg, zhang2021nerfactor, zhang2022modeling, zhu2023i2} have shown realistic view synthesis and consistent estimation of the underlying properties of the scenes. 
Among them, PhySG~\cite{zhang2021physg} and Invrender~\cite{zhang2022modeling} rely heavily on precise geometry, limiting their applicability to specific objects. While IntrinsicNeRF introduces intrinsic decomposition into neural rendering and expands the capabilities of editable novel view synthesis from individual objects to room-scale scenes.


\section{Method}
Given multi-view posed images under unknown illumination of static scenes, we aim to achieve a reliable understanding of the basic properties of the scene, such as reflectance, shading, etc, and to enable real-time editable novel view synthesis. Fig.~\ref{fig: IntrinsicNeRF Framework} outlines the general framework. 


\subsection{Intrinsic Neural Radiance Fields}
\label{Sec: Intrinsic Neural Radiance Fields}
\noindent\textbf{Preliminaries: Intrinsic Decomposition.} Lambertian and grayscale shading assumptions~\cite{fan2018revisiting} are commonly used and introduced to simplify this inverse problem, achieving good approximations of most scenarios. Based on Lambertian assumption, Intrinsic decomposition~\cite{fan2018revisiting} presents an input image $I$ as the pixel-wise product of the illumination-invariant reflectance $R(I)$, and the illumination-varying shading $S(I)$:
	\begin{equation}
	C(I) = R(I) \odot S(I),
	\end{equation}
where $\odot$ is channel-wise multiplication. However, the Lambertian assumption is difficult to be satisfied in realistic scenes, and the intrinsic residual model~\cite{tominaga1994dichromatic, maxwell2008bi} introduces view-independent reflectance and shading with an additional view-dependent residual term $Re(I)$ to model scenes that do not satisfy the Lambertian assumption, such as glossy reflections, metallic materials:
\begin{equation}\label{equa: intrinsic residual}
	C(I) = R(I) \odot S(I) + Re(I).
	\end{equation}
\noindent\textbf{Our representation.} 
IntrinsicNeRF takes the sampled coordinate point $\mathbf{x} = (x,y,z)$ and direction $\mathbf{d}=(\theta, \phi)$ as input, and outputs the view-independent reflectance $r$  and shading $s$, the view-dependent intrinsic residual term $re$ and the volume density $\sigma$ through an MLP network $F_{\Theta}$:
\begin{equation}\label{equa: intrinsicnerf decomposition}
(r, s, re, \sigma) = F_{\Theta}(\mathbf{x}, \mathbf{d}).
\end{equation}
The predicted color c of each spatial point can be obtained by Eq.~\ref{equa: intrinsic residual} and the target color $C(\mathbf{r})$ of camera ray $\mathbf{r}$ is: 
\begin{align}
    \hat{C}(\mathbf{r}) =&
        \sum_{k=1}^{K} \hat{T}_k \, \alpha_k \, c_k \,\,
    \text{and} \,\, \hat{T}_k = \exp \left(-\sum_{k'=1}^{k-1} \sigma_k \delta_k \right),
\end{align}
where $\alpha_k=1-\exp(-\sigma_k \delta_k)$, and $\delta_k$ is the distance between two adjacent sampled points. We follow NeRF's coarse-to-fine training policy and train IntrinsicNeRF from scratch with the photometric loss $L_{pho}$ in NeRF~\cite{mildenhall2020nerf}.


\begin{figure}[h]
  \centering
  \includegraphics[width=\linewidth]{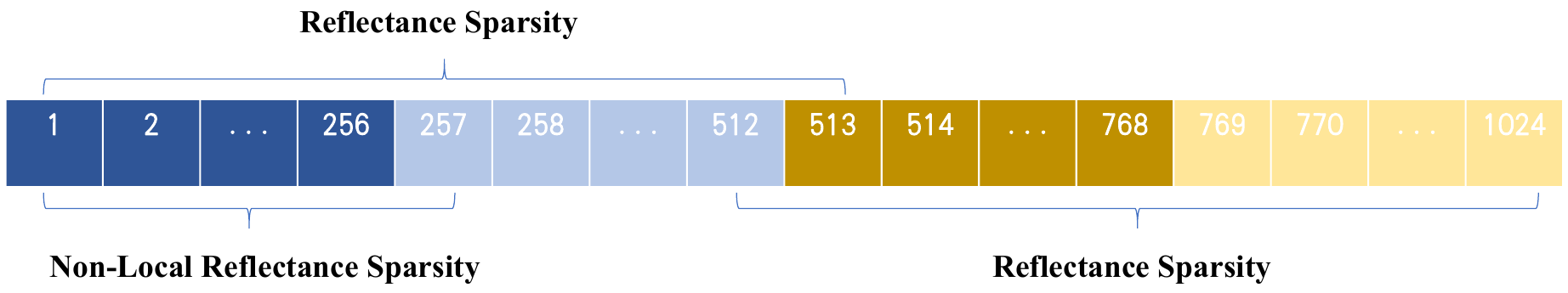}
  \vspace{-1.5em}
  \caption{\textbf{Distance-Aware Point Sampling}. 
  We first randomly sample 512 points, and then randomly sample the remaining 512 points in the eight neighborhoods of each sampled point to construct the unsupervised constraint term for the intrinsic decomposition.
  }
  \label{fig: Distance-Aware Point Sampling}
\end{figure} 

\noindent\textbf{Distance-Aware Point Sampling.}
\label{Sec: Distance-Aware Point Sampling}
NeRF~\cite{mildenhall2020nerf} randomly samples batches of camera rays from the image pixel set (roughly 1024 points) in each optimization, where these points are random, and no relationship is established between them.  
It is not applicable in IntrinsicNeRF, for the introduction of ill-posed intrinsic decomposition into NeRF makes the whole optimization process stochastic, as shown in Fig.~\ref{fig: Quality Results on Blender Dataset} (Baseline column). 
To this end, we make a sophisticated design of the sampling policy (see Fig.~\ref{fig: Distance-Aware Point Sampling}) which helps to construct intrinsic prior constraints (see Sec.~\ref{Sec: Unsupervised Priors}), and the network can be trained unsupervised. 

\subsection{Unsupervised Prior Guidance}
\label{Sec: Unsupervised Priors}
Following intrinsic decomposition works~\cite{meka2016live}, we adopt the grayscale shading assumption to simplify this inverse problem, so that the shading layer is single-channel and the reflectance chromaticity of the image $I$ is approximated to $c(\mathbf{x}) = I(\mathbf{x})/|I(\mathbf{x})|$. We define the chromaticity similarity weight $\omega_{cs} (\mathbf{x}, \mathbf{y})$~\cite{meka2016live} that is associated with many priors:
\begin{align}
    \omega_{cs(\mathbf{x}, \mathbf{y})} & = \exp (- \alpha_{cs} {\left\lVert c(\mathbf{x}) - c(\mathbf{y}) \right\rVert}_2^2),
\end{align}
where $\mathbf{x}$ and $\mathbf{y}$ are the image pixel coordinates. Coefficient $\alpha_{cs}$ = 60 produces the best decomposition results.

\noindent\textbf{Chromaticity Prior.} 
Due to the residual term, the chromaticity of the unknown reflectance $R$ and the input image $I$ are not the same. We want them to be as close as possible:
\begin{align}
    L_{chrom(\mathbf{x})} & = 
    {\left\lVert c_r(\mathbf{x}) - c(\mathbf{x}) \right\rVert}_2^2,
\end{align}
where $c$ and $c_r$ are the chromaticity of the input sample points and the sampled points' reflectance, respectively.

\noindent\textbf{Reflectance Sparsity.}
\label{Reflectance Sparsity}
Two pixels that are similar in spatial location and chromaticity, have converging reflectance $r$, which leads to reflectance sparsity. Following ~\cite{meka2016live}, we minimize the reflectance gradients magnitude independently: 
\begin{align}
    L_{reflect(\mathbf{x})} & =  
    \sum_{\mathbf{y}\in \mathcal{N}(\mathbf{x})} \omega_{cs(\mathbf{x}, \mathbf{y})} {\left\lVert r(\mathbf{x}) - r(\mathbf{y}) \right\rVert}_2^2,
\end{align}
where $\mathcal{N}(\mathbf{x})$ is the neighbourhood of pixel $\mathbf{x}$. 
Specifically, in IntrinsicNeRF, the sampled points in the first half will be adjacent to the second half, shown in Fig.~\ref{fig: Distance-Aware Point Sampling}.

\noindent\textbf{Non-Local Reflectance Sparsity.}
\label{Spatio-Temporal Reflectance Consistency Prior} In natural and man-made scenes, two distant spatial points may also have the same reflectance, such as a wall and floor that occupy a larger image area, which requires non-local reflectance sparsity. In the sampling of IntrinsicNeRF, the first half of the points are randomly sampled, so the distance between any two points can be very far. We simply bisect the first half of the points and construct a non-local reflectance sparsity constraint (following ~\cite{meka2016live}) on the points in the first 1/4 segment and the corresponding points in the next 1/4 segment:
\begin{align}
    L_{non-local(\mathbf{x})} & = \sum_{\mathbf{y}\in \mathcal{F}(\mathbf{x})} \omega_{cs(\mathbf{x},\mathbf{y})} {\left\lVert r(\mathbf{x}) - r(\mathbf{y}) \right\rVert}_2^2,
\end{align}
where $\mathcal{F}(\mathbf{x})$ is the farhood of pixel $\mathbf{x}$. Note that the weight of this constraint is smaller than the reflectance sparsity's.

\noindent\textbf{Shading Smoothness.}
\label{Shading Smoothness}
Natural objects usually have smooth surfaces and the shading variance is expected to be smooth~\cite{meka2016live}. Moreover, neighboring pixels with different chromaticities, represent a reflectance edge, so we strongly enforce the shading smoothness:
\begin{align}
    L_{shade(\mathbf{x})} &= \sum_{\mathbf{y}\in \mathcal{N}(\mathbf{x})}{\left\lVert c(\mathbf{x})-c(\mathbf{y}) \right\rVert}_2^2 {\left\lVert s(\mathbf{x}) - s(\mathbf{y}) \right\rVert}_2^2.
    \vspace{-1.2em}
\end{align}

\begin{figure}[hb]
\centering
\includegraphics[width=\linewidth]{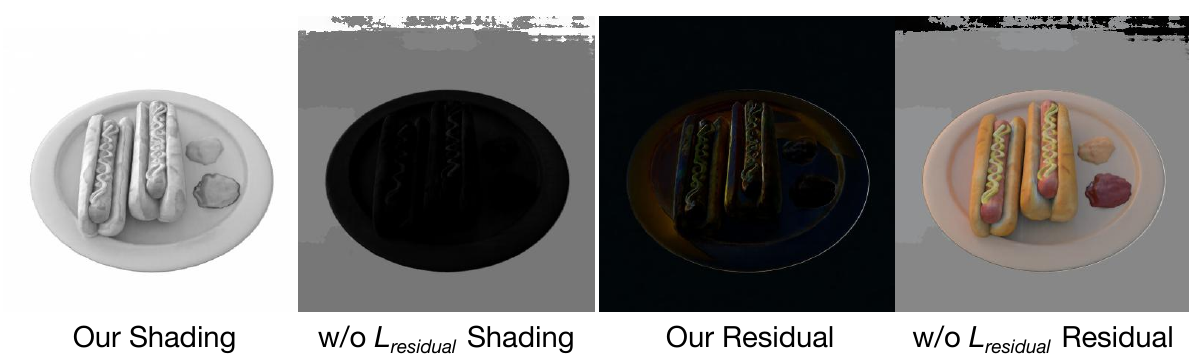}
\vspace{-1.5em}
\caption{\textbf{Ablation Study of $L_{residual}$.} Without the residual constraint, the reflectance gets worse (see Tab.~\ref{tab:ablation}), and the shading and residual are exceptionally unfavorable.}
\label{fig:w/o residual}
\end{figure}

\noindent\textbf{Intrinsic Residual Constraints.} Since diffuse light generally dominates the scene, we want the image content to be recovered by reflectance and shading as much as possible. 
This prevents extreme cases when $R$ and $S$ both converge to zero and $Re = I$, which would destroy the efficacy of the previous constraints and fall into catastrophic results (see Fig.~\ref{fig:w/o residual}). 
We set this constraint as follows:
\begin{align}
    L_{residual(\mathbf{x})} & = 
    {\left\lVert re(\mathbf{x})  \right\rVert}_2^2.
\end{align}
The weight is set higher early, so $R(I) \odot S(I)$ is close to the target image $I$ and then dropped lately. As the output of $R$ and $S$ is stable, $R_e$ can represent the view-dependent components, such as glossy reflections.

\noindent\textbf{Intensity Prior.}
\label{Intensity Prior} The previous constraints on reflectance and shading only consider the relative relationship between two pixels. The absolute magnitude of $R$ and $S$ is required to prevent them from falling into certain extremes during optimization. The intensity of the unknown reflectance image $R$ and the input image $I$ should be close: 
\begin{align}
    L_{intensity(\mathbf{x})} & = 
    {\left\lVert i_r(\mathbf{x}) - i(\mathbf{x}) \right\rVert}_2^2, 
\end{align}
where $i$ and $i_r$ are the average intensities of the sampled points $\mathbf{x}$ of the input image and reflectance $r$. The weight of this constraint is set higher early and then reduced.

\begin{figure}[h]
  \centering
  \includegraphics[width=\linewidth]{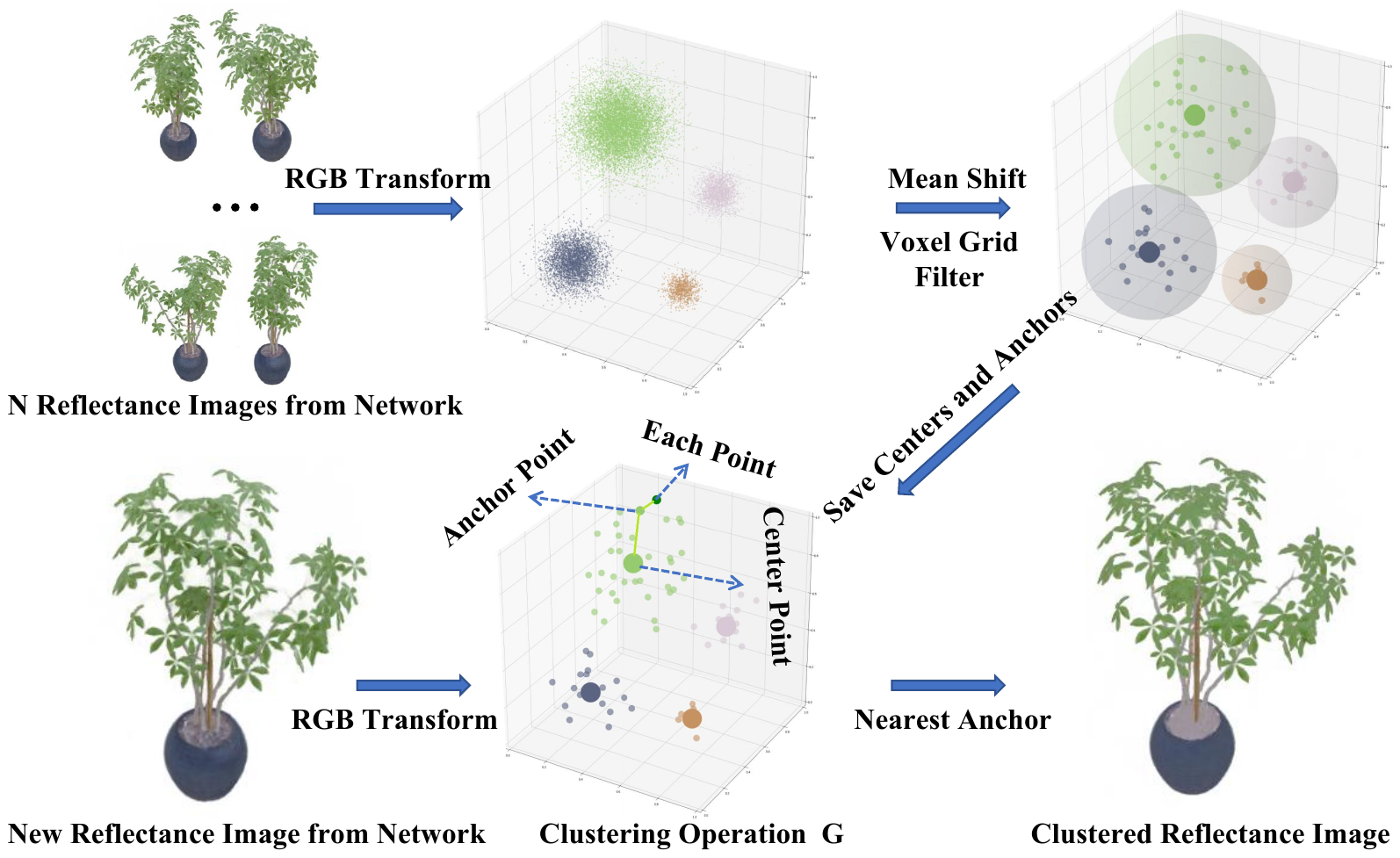}
  \vspace{-1.5em}
  \caption{\textbf{Adaptive Reflectance Iterative Clustering Method}. The color of the reflectance pixels is first converted and then clustered with the Mean Shift algorithm. The voxel grid filter is performed to accelerate the processing of the fast approximation of clustering operation $G$.
  }
  \label{fig: Reflectance Clustering with Mean Shift}
  \vspace{-1.1em}
\end{figure}

\subsection{Adaptive Reflectance Iterative Clustering}
\label{Sec: Adaptive Reflectance Clustering with Mean Shift}
Although reflectance sparsity makes sense to some extent, there still remain inconsistencies of similar reflectance regions (see Fig.~\ref{fig: Quality Results on Blender Dataset} Baseline+w/prior),  therefore we propose an adaptive reflectance iterative clustering method by constructing a continuously updated clustering operation $G$, which maps similar reflectance colors $r(\mathbf{x})$ to the same target reflectance color $r_{cluster}(\mathbf{x})$ by adding a clustering constraint during the optimization of the network:
\begin{align}
    L_{cluster(\mathbf{x})} & = 
    {\left\lVert r_{cluster}(\mathbf{x}) - r(\mathbf{x}) \right\rVert}_2^2.
\end{align}

Next, we elucidate the detail of the clustering method.

\noindent\textbf{RGB Transform.} During the training of the network, we infer the reflectance $r$, shading $s$, and residual term $re$ of multi-view posed images after every 10K iterations. Refer to IIW~\cite{bell2014intrinsic}, we take out all pixels of all $r$ components and convert their colors to better cluster reflectances (pixel intensity, red chromaticity, green chromaticity~\cite{meka2016live}):
\begin{align}
    \mathbf{\textit{\textbf{f}} ([r, g, b])} = \mathbf{[\beta \frac{r+g+b}{3},\frac{r}{r+g+b},\frac{g}{r+g+b}]},
\end{align}
where  $\beta$ is set as 0.5~\cite{bell2014intrinsic} in our experiment. The RGB transformation helps reduce the effect of intensity differences on the clustering, making the clustering more focused on the similarity of chromaticity between two pixels. The transformed RGB space is considered as \textit{\textbf{f}} space.

\noindent\textbf{Mean Shift.} Unlike existing methods~\cite{meka2016live} using K-Means clustering to specify K clustering categories, we instead cluster all the pixel points $P$ every 10K iterations with a Mean Shift clustering algorithm to adaptively determine the number of reflectance classes in the scene, for we do not know the reflectance class number. 

\noindent\textbf{Clustering Operation $\textit{G}$.} 
After Mean Shift clustering, we get a set of clustered centers, and a classification label for each pixel point in $P$. During each training iteration, it is unrealistic to cluster the reflectance of each rendered pixel because the clustering is time-consuming. So we define a fast approximation clustering operation $G$: for an RGB value of reflectance, it considers the category of the nearest point in $P$ as its cluster category and set the value of the category center as the target clustered reflectance $r_{cluster}(\mathbf{x})$. When calculating the clustering loss, we only use the Clustering Operation G, shown in Fig.~\ref{fig: Reflectance Clustering with Mean Shift}.

\noindent\textbf{Voxel Grid Filter.} Since there are plenty of points $P$ in the \textit{\textbf{f}} space and most of them are clustered in very small regions due to reflectance sparsity, rather than finding the nearest neighbors in all points, we perform voxel grid filter (voxel size is $0.01$) on the points $P$ in the \textit{\textbf{f}} space, and the filtered points are regarded as anchor points. The clustering operation $G$ therefore only needs to search the closest anchor point, and the anchor points are only been updated every 10K iterations by Mean-Shift.  

\noindent\textbf{Optimization.} During the network optimization, the weight of the clustering loss $L_{cluster(\mathbf{x})}$ and the bandwidth parameter in the mean shift algorithm are gradually increased with the number of iterations (the larger the bandwidth is, the smaller the number of mean-shift clustering categories is). That is because, in the early stage of network optimization, the inferred reflectance $r$ is not reliable and needs lower weight. While in the later stage, a higher weight can lead the output of the network to converge toward the effect of clustering, making the reflectance $r$ before and after clustering indistinguishable.

\begin{figure}[t!]
    \centering
    \includegraphics[width=1.0\linewidth, trim={0 0 0 0}, clip]{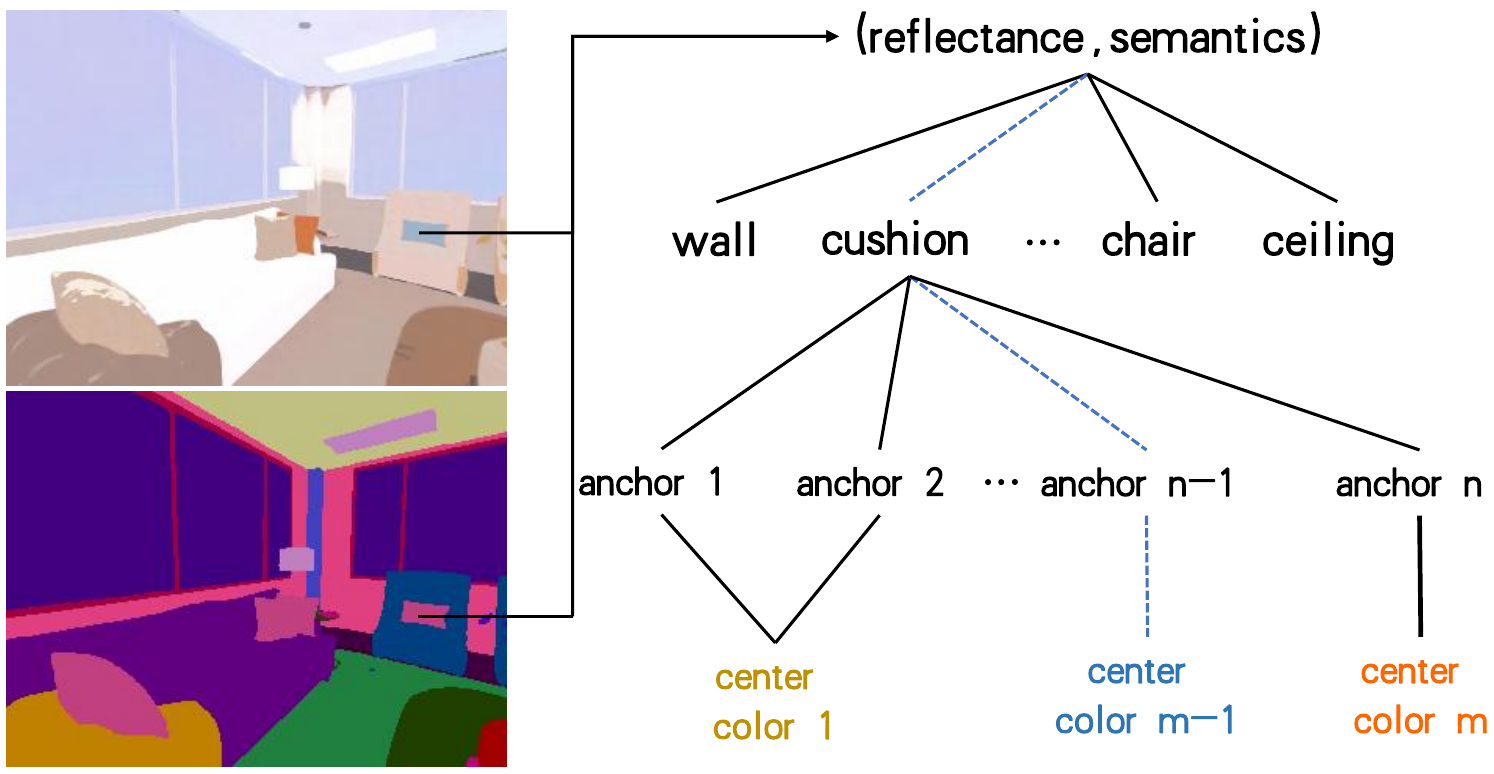}
    \vspace{-1.5em}
  \caption{\textbf{Hierarchical Clustering and Indexing Method.} 
  }
  \label{fig: Hierarchical Clustering and Indexing}
  \vspace{-1.1em}
\end{figure}

\subsection{Hierarchical Clustering and Indexing}
\label{Sec: Hierarchical Reflectance Clustering}
\begin{figure*}[h] 
  \centering
  \includegraphics[width=\linewidth]{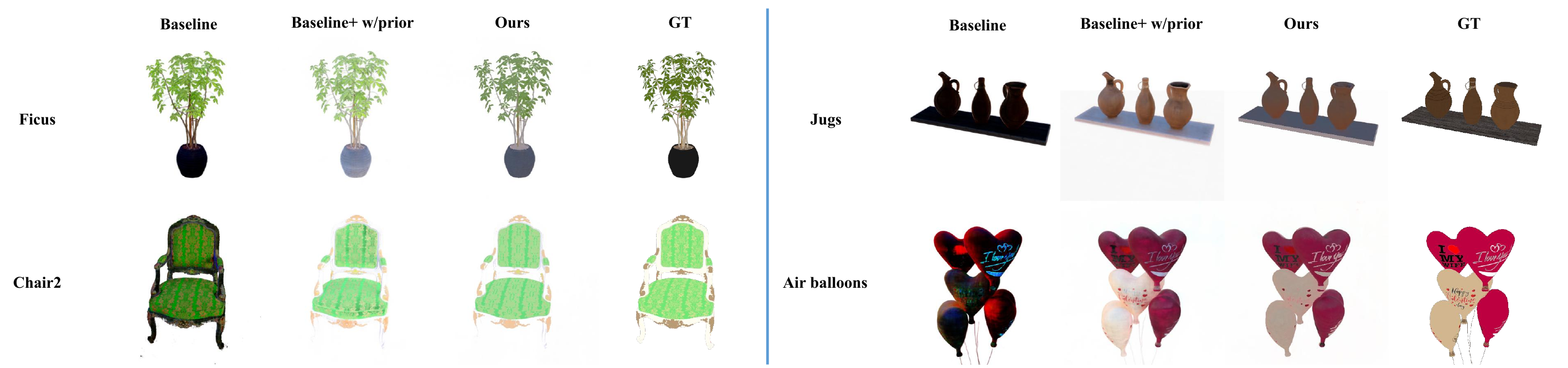}
  \vspace{-2em}
  \caption{\textbf{Ablation study of Reflectance Estimation Sample on the Blender Object Dataset.} Left: our dataset, right: Invrender dataset. The reflectance estimation of the baseline method is stochastic and unstable, while the intrinsic prior makes the optimization of the network traceable. Our final model achieves more plausible reflectance results. 
  }
  \label{fig: Quality Results on Blender Dataset}
\end{figure*}

\begin{figure*}[h]
  \centering
\includegraphics[width=\linewidth]{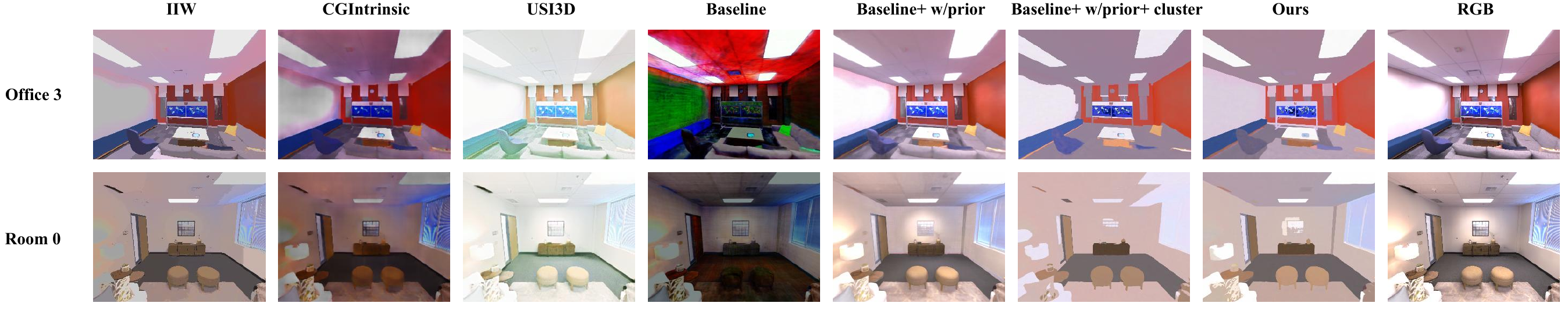}
\vspace{-1.5em}
  \caption{\textbf{Qualitative Reflectance  Comparison Samples with Previous Methods on the Replica Scene Dataset.} Experiments demonstrate the progressive facilitation effect of our different variants. Compared with previous methods, our final method achieves more plausible and consistent reflectance estimation results, retaining the boundaries of objects. 
}
  \label{fig: Comparisons with Previous work on Replica Scene}
\end{figure*}

The adaptive reflectance iterative clustering method can handle object-level scenes well, shown in Fig.~\ref{fig: Quality Results on Blender Dataset} (Ours). However, when the reflectance in the scene is complex and similar, plenty of different instances with similar reflectance in room-scale scenes may be incorrectly clustered, shown in Fig.~\ref{fig: Comparisons with Previous work on Replica Scene} (w/prior+cluster). So we propose a semantic-aware reflectance sparsity constraint, where only pixels with the same semantic label will be computed, thus significantly improving the quality of reflectance. Inspired by~\cite{zhi2021place}, we extend IntrinsicNeRF to jointly encode appearance, geometry, and semantics by adding a segmentation renderer to the original IntrinsicNeRF. Specially, we use a view-invariant MLP function $sl = F_{\Theta}(\mathbf{x})$ to map a spatial coordinate $\mathbf{x}$ to semantic label and use the semantic loss $L_{sem}$ in~\cite{zhi2021place}.



\begin{table*}[t]
\small
\begin{center}

\resizebox{\textwidth}{18mm}{
\begin{tabular}{ c c c c c c c c c c c c c c c c c c c}
\toprule
& \multicolumn{5}{c}{Reflectance (Invrender dataset)} & \multicolumn{3}{c}{View Synthesis (Invrender dataset)} &
 \multicolumn{5}{c}{Reflectance (our dataset)} & \multicolumn{3}{c}{View Synthesis (our dataset)}\\
 \cmidrule(lr){1-1}\cmidrule(lr){2-6}\cmidrule(lr){7-9}\cmidrule(lr){10-14} \cmidrule(lr){15-17}
 Method 
 & PSNR $\uparrow$ & SSIM $\uparrow$ & LPIPS $\downarrow$ & MSE $\downarrow$ & LMSE $\downarrow$ 
 & PSNR $\uparrow$ & SSIM $\uparrow$ & LPIPS $\downarrow$ 
 & PSNR $\uparrow$ & SSIM $\uparrow$ & LPIPS $\downarrow$ & MSE $\downarrow$ & LMSE $\downarrow$ 
 & PSNR $\uparrow$ & SSIM $\uparrow$ & LPIPS $\downarrow$ \\
\midrule

IIW\cite{bell2014intrinsic} & 22.0284 &
0.9307  & \underline{0.0847}  & 0.0099  &
\textbf{0.0120} & - & - & 
 -  & 20.5299  & 0.9079  & 
0.1131  & 0.0102 & 0.0727 &
 -  &  -  &  -  \\ 

CGIntrinsic\cite{li2018cgintrinsics} & 20.1583 &
0.9209 &0.0996&0.0129&
\underline{0.0141} & - & - & 
- &18.3542 &0.8999&
0.1229 & 0.0156 & 0.0659 &
 -  &  -  &  - \\ 

USI3D~\cite{liu2020unsupervised} & 20.7571 &
0.9267  & 0.0887 & 0.0079 & 
0.0149 & - & - & 
- & 19.1489 & 0.9115 & 
\underline{0.1070} & 0.0135 & 0.0524 &
 -  &  -  &  - \\
 
 \midrule



NeRFactor\cite{zhang2021nerfactor}  & 19.9167 & 
0.9156 & 0.1354 & 0.0059 & 0.0210 & 23.0133 & 0.9277 & 0.0822 & 21.4440 & 0.9170 &
 0.1055  &  \textbf{0.0063}  &  \textbf{0.0444}   &
 20.6880  &  0.8733  &  0.1185    \\


PhySG\cite{zhang2021physg} & 23.3748  & 0.9231
 & 0.1092 & 0.0034 & 0.0396
 &25.4225  & 0.9388  & 0.0804
 & - & - &
-  &  -  &  -   &
-  &  -  &  -    \\ 

Invrender~\cite{zhang2022modeling} & \textbf{26.3078} &
\textbf{0.9380} & \textbf{0.0572} & \underline{0.0022} & 
0.0226 & 29.3870 & 0.9522 & 
0.0505 & - & - &
 -  &  -  &  -   &
 -  &  -  &  -    \\
\midrule

Baseline & 16.3209 &
0.8637 & 0.1301 & 0.0254 & 
0.1955 & \textbf{34.0036} & \textbf{0.9670} & 
\textbf{0.0252} & 14.8572 & 0.8397 & 
0.1738 & 0.0451 & 0.1849 & 
\textbf{28.2604} & \textbf{0.9383} & \textbf{0.0339} \\

Baseline + w/ prior. & 21.7370 &
0.9278 & 0.1086 & 0.0055 &
0.0186 & \underline{33.4909} & \underline{0.9638} & 
\underline{0.0304} & \underline{20.9646} & \underline{0.9140} & 
0.1216 & 0.0095 & 0.0538 & 
\underline{28.0633} & \underline{0.9370} & \underline{0.0369} \\

Ours & \underline{24.2642} &
\underline{0.9371} & 0.0880 & \textbf{0.0021} & 
0.0173 & 33.4967 & 0.9630 & 
0.0306 & \textbf{22.5677} & \textbf{0.9267} & 
\textbf{0.0975} & \underline{0.0066} & \underline{0.0474} &
27.9494 & 0.9357 & 0.0372 \\

\bottomrule
\end{tabular}}
\vspace{-0.5em}
\caption{\textbf{Quantitative Results of the Blender Object Dataset.}
For reflectance estimation, IntrinsicNeRF achieved the best results on our dataset and ranked 2nd on the Invrender dataset. For novel view synthesis, IntrinsicNeRF achieved the best performance on both datasets, while Invrender~\cite{zhang2022modeling} and PhySG~\cite{zhang2021physg} require good geometric prerequisites, which makes them fail on our dataset. Moreover, intrinsic decomposition methods can not perform novel view synthesis. 
- means failure.
}
\label{tab:quantitative object}
\end{center}
\end{table*}

\begin{table*}[t]
\small
\begin{center}
\vspace{-1.8em}
\resizebox{\linewidth}{!}
    {
\begin{tabular}{l|ccccccccc}
 \hline
 \diagbox{Metric}{Method} & w/o $L_{chrom}$ & w/o $L_{reflect}$ & w/o $L_{non-local}$  & w/o $L_{shade}$ & w/o $L_{cluster}$ & w/o $L_{residual}$  & w/o $L_{intensity}$  & w/o all prior & Ours \\
 \hline
 PSNR $\uparrow$ &22.0243 &22.4955 &23.3032 &22.9874  &21.3508 &21.1288 &18.7466 &15.5891 & \textbf{23.4160}\\
 MSE $\downarrow$ &0.0067 &0.0060 &0.0044 &0.0048  &0.0075 &0.0074 &0.0172	&0.0352 &\textbf{0.0043}\\
 LMSE $\downarrow$ &0.0392 &0.0378 &0.0323 &0.0338  &0.0362 &0.0387 &0.0339 &0.1902 &\textbf{0.0323}\\
 \hline
 \end{tabular}
    }
    \vspace{-0.5em}
    \caption{\textbf{Ablation Studies of Each Loss Constraints for Reflectance Estimation on the Blender Object Dataset.} }
    \label{tab:ablation}
\end{center}
\vspace{-2em}
\end{table*}

\label{Sec: Hierarchical Clustering and Indexing}
Depending on the semantic labels of each pixel, the pixel set $P$ can be divided into multiple subsets $\{\mathbf{P_1, P_2, ..., P_N}\}$, where N is the number of semantic categories. Then we can construct N clustering operations $\{\mathbf{G_1, G_2, ..., G_N}\}$ as Sec.~\ref{Sec: Adaptive Reflectance Clustering with Mean Shift}. 
The hierarchical clustering operation takes the reflectance RGB value of each pixel and the corresponding semantic label as input and outputs the result of the clustering operation for the pixel under the semantic label.
Such a hierarchical clustering method allows the clustered information of each pixel to be stored in a tree structure, shown in Fig.~\ref{fig: Hierarchical Clustering and Indexing}, which can be indexed quickly. 

\begin{figure*}[h]
  \centering
  \includegraphics[width=\linewidth]{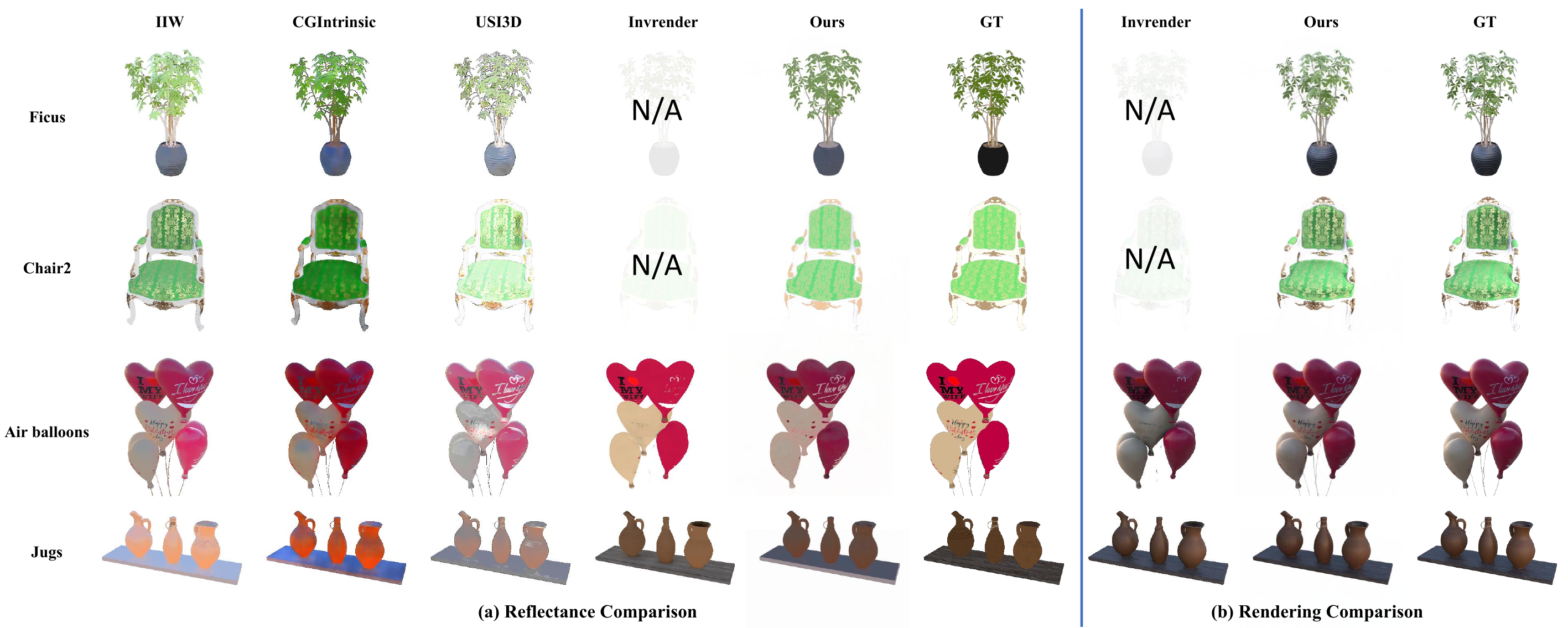}
  \vspace{-2em}
  \caption{\textbf{Qualitative Comparison Results of Reflectance and Rendering
  on the Blender Object Dataset.} The top 2 rows represent our samples and the bottom 2 rows are the Invrender samples. Our method can perform reflectance estimation and novel view synthesis on both datasets well, while Invrender~\cite{zhang2022modeling} fails to do that on our dataset. N/A means failure. 
  }
  \label{fig: Comparisons with Previous work on Blender Datasets}
  \vspace{-1.1em}
\end{figure*}

\subsection{Implementation Details}
\label{sec.supp.Implementation Details}
We implement IntrinsicNeRF on the top of SemanticNeRF~\cite{zhi2021place} with additional three FC layers for intrinsic components which have 128 neurons. The network is optimized with photometric loss, semantic loss, unsupervised prior constraints, and clustering loss jointly.
The final loss is:
\begin{align}
\vspace{-0.2em}
L_{final} & = \lambda_{pho} L_{pho} +  \lambda_{sem} L_{sem} + \lambda_{chrom} L_{chrom}  \nonumber\\ & + \lambda_{reflect} L_{reflect} 
 + \lambda_{non-local} L_{non-local}  \nonumber\\
 & + \lambda_{shade} L_{shade} \nonumber + \lambda_{cluster} L_{cluster} \\
& + \lambda_{residual} L_{residual} 
+ \lambda_{intensity} L_{intensity}.
\end{align}
Here, $\lambda_{pho}=1$, $\lambda_{sem}=0.04$, $\lambda_{chrom}=1$, $\lambda_{reflect}=0.01$, $\lambda_{non-local}=0.005$ and $\lambda_{shade}=1$. While $\lambda_{cluster} = 10^{-2(1-iter/200K)}$, it exponentially increases from 0.01 to 1 every 10K iterations. We set $\lambda_{residual}=1$ in the early 100K iterations and dropped to 0.02 in the later iterations. The $\lambda_{intensity}$ is set to 0.1 in the first 50K iterations and then set to 0.01. The batch size of the rays is 1024. The Adam~\cite{kingma2014adam} optimizer is used with a learning rate of 5e-4 for 200K iterations. Tab.~\ref{tab:time} shows the acceptable clustering and the total training time of our method.

\begin{table}[!ht]
\centering
  \resizebox{0.4\textwidth}{!}{
  \begin{tabular}{lcccc}
    \toprule
    Dataset & Method & Training & Test & Clustering \\
    \midrule
    Blender Object & NeRFactor~\cite{zhang2021nerfactor} & 5.7d & 5.65s & - \\
    Blender Object & PhySG~\cite{zhang2021physg} & 5.2h & 2.92s & - \\
    Blender Object & InvRender~\cite{zhang2022modeling} & 11.4h & 14.25s & - \\
    \midrule
    Blender Object & NeRF~\cite{mildenhall2020nerf} & 5.4h & 4.59s & - \\
    Blender Object & Ours & 6.5h & 5.35s & 39s \\
    \midrule
    Replica Scene & Semantic-NeRF~\cite{zhi2021place} & 13.5h & 2.50s & - \\
    Replica Scene & Ours & 17.5h & 2.79s & 220s \\
    \bottomrule
  \end{tabular}}
  \vspace{-0.5em}
  \caption{\textbf{Time Comparison.} We show the total training time, the average synthesis time of each frame, and the average clustering time of our method, where \textbf{the clustering is performed every 10K training iterations.} All run on a single RTX3090.}
  \label{tab:time}
  \vspace{-1.2em}
\end{table}

\section{Experiments}
\label{Sec:Experiments}
We first make qualitative and quantitative comparisons of IntrinsicNeRF with traditional optimization-based~\cite{bell2014intrinsic} and learning-based~\cite{li2018cgintrinsics, liu2020unsupervised} intrinsic decomposition methods, and neural rendering methods~\cite{zhang2021nerfactor, zhang2021physg, zhang2022modeling} combined with inverse rendering on synthetic object dataset in Sec.~\ref{Sec: IntrinsicNeRF for Blender Object Datasets}. 
Then we only compare qualitative results on synthetic 
scenes (e.g. Replica~\cite{replica19arxiv}) in Sec.~\ref{Sec: IntrinsicNeRF for Scene Datasets}, due to the lack of ground-truth labels.
Finally, we perform ablation studies in Sec.~\ref{Sec:Ablation Studies} to analyze the design of our framework and demonstrate its applicability in Sec.~\ref{Sec:Applications} to both synthetic and real-world data. 

\subsection{Dataset} 
\label{Sec: Dataset}

\noindent\textbf{Synthetic Data.} We collect 8 Blender Object dataset (4 from Invrender~\cite{zhang2022modeling}, and 4 from NeRF~\cite{mildenhall2020nerf}) and 8 Replica Scene dataset. 
\label{Sec: Blender Dataset}
The Invrender dataset contains Hotdogs, Jugs, Chair, and Air balloons, and each dataset is rendered by Blender Cycles~\cite{blender} with their masks, reflectance, and roughness maps. The NeRF dataset contains 4 objects (Lego, Drums, Ficus, and Chair2) that maintain complex geometry and realistic non-Lambertian materials. Note that some environment lighting maps in NeRF's open-source blender model were missing, we search for some environment maps that look as realistic as possible and re-render the new image to match NeRF's settings. We regard this dataset as our dataset. The image resolution is set as 400×400.

\label{Sec: Replica Scene} 
Generated by Semantic-NeRF~\cite{zhi2021place}, each Replica Scene~\cite{replica19arxiv} of rooms and offices consists of RGB images, depth maps, and semantic labels at resolution 320x240 from randomly generated 6-DOF trajectories. It contains challenging illumination effects, such as glossy reflections. 


\noindent\textbf{Real-world Data.} 
We selected 4 real data of natural scenes (Orchids, Flowers, Horns, and Ferns) at 504x378 resolution from LLFF~\cite{mildenhall2019llff} to demonstrate the generalization ability of our method in real-world lighting and reflection and its applicability such as recoloring, illumination variation.
\subsection{Comparison on the Blender Object Dataset}
\label{Sec: IntrinsicNeRF for Blender Object Datasets}
We exploit Peak Signal-to-Noise Ratio (PSNR), Structural Similarity Index Measure (SSIM), Learned Perceptual Image Patch Similarity (LPIPS)~\cite{zhang_unreasonable_2018}, Mean Squared Error (MSE) and Local Mean Squared Error (LMSE) as reflectance evaluation metrics. We do not evaluate the shading quantitatively because different methods model lighting differently, and we cannot get the ground-truth shading of our model in Blender (Eq.~\ref{equa: intrinsic residual}). In contrast, reflectance is a common output and has ground-truth values, so we focus on the evaluation of reflectance. The view synthesis evaluation metrics are PSNR, SSIM, and LPIPS. 

\begin{table}
\centering
  \begin{tabular}{cccl}
    \toprule
    Method & PSNR $\uparrow$ & SSIM $\uparrow$ & LPIPS $\downarrow$ \\
    \midrule
    NeRF~\cite{mildenhall2020nerf} & 31.0838  & 0.9525 & 0.0302 \\
    Ours & 30.7230 & 0.9494 & 0.0339 \\
    \bottomrule
  \end{tabular}
  \vspace{-0.5em}
  \caption{\textbf{Comparable Quantitative Results for Novel View Synthesis on the Blender Object Dataset.} 
  }
  \label{tab:compared with NeRF}
  \vspace{-1em}
\end{table}

\begin{table}[t]
\centering
  \begin{tabular}{ccccl}
    \toprule
    Method &PSNR $\uparrow$ &SSIM $\uparrow$ &LPIPS $\downarrow$ &mIoU $\uparrow$ \\
    \midrule
   ~\cite{zhi2021place} &30.9770 &0.8955 &0.1066 &0.9725 \\
    Ours &30.7044 &0.8908 &0.1140 &0.9702 \\
    \bottomrule
  \end{tabular}
  \vspace{-0.5em}
  \caption{\textbf{Comparable Results for View Synthesis and Semantic Segmentation on the Replica Scene Dataset.} 
  }
  \label{tab:scene iou nvs}
  \vspace{-1.5em}
\end{table}

We compare IntrinsicNeRF with the following methods:
IIW~\cite{bell2014intrinsic} is a classic intrinsic decomposition method that does not require training. CGIntrinsic~\cite{li2018cgintrinsics} is a learning method with good generalization trained on large-scale datasets, and USI3D~\cite{liu2020unsupervised} is another state-of-the-art unsupervised learning method, and we use their pre-trained models. We do not choose IRISformer~\cite{zhu2022irisformer} and intrinsic video decomposition methods~\cite{meka2016live, meka2021real}, because their codes are not available.
NeRFactor~\cite{zhang2021nerfactor}, PhySG~\cite{zhang2021physg}, and InvRender~\cite{zhang2022modeling} are the neural rendering methods, and we have retrained them in the same setting for fair comparisons. 
Tab.~\ref{tab:quantitative object} shows our method achieves the best results on our dataset and ranked 2nd on the Invrender dataset for reflectance estimation. Compared with single-view intrinsic decomposition methods, our method yields more consistent and plausible decomposition results, even in challenging object scenes, such as Chair2, and Ficus.
As for view synthesis, IntrinsicNeRF achieves the best performance on both datasets, while Invrender~\cite{zhang2022modeling} and PhySG~\cite{zhang2021physg} require good geometric prerequisites using IDR method~\cite{yariv2020idr}, which makes them fail on our dataset, as shown in Fig.~\ref{fig: Comparisons with Previous work on Blender Datasets}. Moreover, traditional intrinsic decomposition methods can not perform novel view synthesis.
Tab.~\ref{tab:compared with NeRF} shows IntrinsicNeRF achieves comparable novel view synthesis results with NeRF~\cite{mildenhall2020nerf} while giving the power of modeling the intrinsic components of scenes.
\subsection{Comparison on the Replica Scene Dataset}
\label{Sec: IntrinsicNeRF for Scene Datasets}
We only compare qualitative results with intrinsic decomposition methods~\cite{liu2020unsupervised,bell2014intrinsic,li2018cgintrinsics} on the Replica Scene dataset in reflectance estimation, because we cannot obtain the ground truth of reflectance. Fig.~\ref{fig: Comparisons with Previous work on Replica Scene} shows that we can obtain more plausible results than other intrinsic decomposition methods, and maintain consistent reflectance estimation for multi-view images in the supplementary material.
Moreover, our method obtains comparable results with Semantic-NeRF~\cite{zhi2021place} in novel view synthesis and semantic segmentation (the metric is mIOU), shown in Tab.~\ref{tab:scene iou nvs}, 
and we give Semantic-NeRF the ability to model the intrinsic properties of the scene (Fig.~\ref{fig:teaser}).
While PhySG~\cite{zhang2021physg} and Invrender~\cite{ zhang2022modeling} fail to do that in room-scale scenes. 

\subsection{Ablation Studies}
\label{Sec:Ablation Studies}
We perform ablation studies to analyze three components of our method that primarily affect the intrinsic decomposition quality. The baseline method is the NeRF~\cite{mildenhall2020nerf} variant with intrinsic neural radiance fields, using the proposed distance-aware point sampling policy. Tab.~\ref{tab:quantitative object} shows that the introduction of the intrinsic prior and iterative clustering leads to more accurate reflectance estimation, with a slight decrease in the accuracy of the novel view synthesis. Fig.~\ref{fig: Quality Results on Blender Dataset} show that the reflectance estimated by the baseline method is more stochastic and unstable. While adding the intrinsic prior, the network output is plausible. The adaptive reflectance iterative clustering method can make the reflectance regions of the same material cluster together. The average results of the Blender Object dataset in Tab.~\ref{tab:ablation} show the effectiveness of our method and the necessity of each loss.
However, reflectance clustering may lose some distinguishable boundaries in room-scale scenes such as Replica Scene, for complex and similar reflectance may be clustered incorrectly. Whereas the hierarchical clustering method with semantic constraints can retain the boundaries and still yields more plausible results, as shown in Fig.~\ref{fig: Comparisons with Previous work on Replica Scene}. 
See more qualitative comparison results of IntrinsicNeRF variants in different scenarios in the supplementary material.


\begin{table}[hb]
\vspace{-0.5em}
\centering
\footnotesize
\setlength{\tabcolsep}{1.2mm}
\begin{tabular}{l|cccc}
 \hline
 \diagbox{Metric}{Method} & IIW~\cite{bell2014intrinsic}  & CGIntrinsic~\cite{li2018cgintrinsics}  & USI3D~\cite{liu2020unsupervised} & Ours \\
 \hline
 PSNR $\uparrow$ &16.1219 &17.9740 &12.8545 & 15.4692 \\
 MSE $\downarrow$ &0.0263 &0.0159 &0.0655 &0.0345\\
 LMSE $\downarrow$ &0.1498 &0.1342 &0.1675 &0.1415\\
 \hline
\end{tabular}
\vspace{-.5em}
\caption{Reflectance Comparison on the EDEN dataset.}
\label{tab:EDEN}
\end{table}

\begin{figure}[hb]
\centering
\vspace{-2em}
\includegraphics[width=\linewidth]{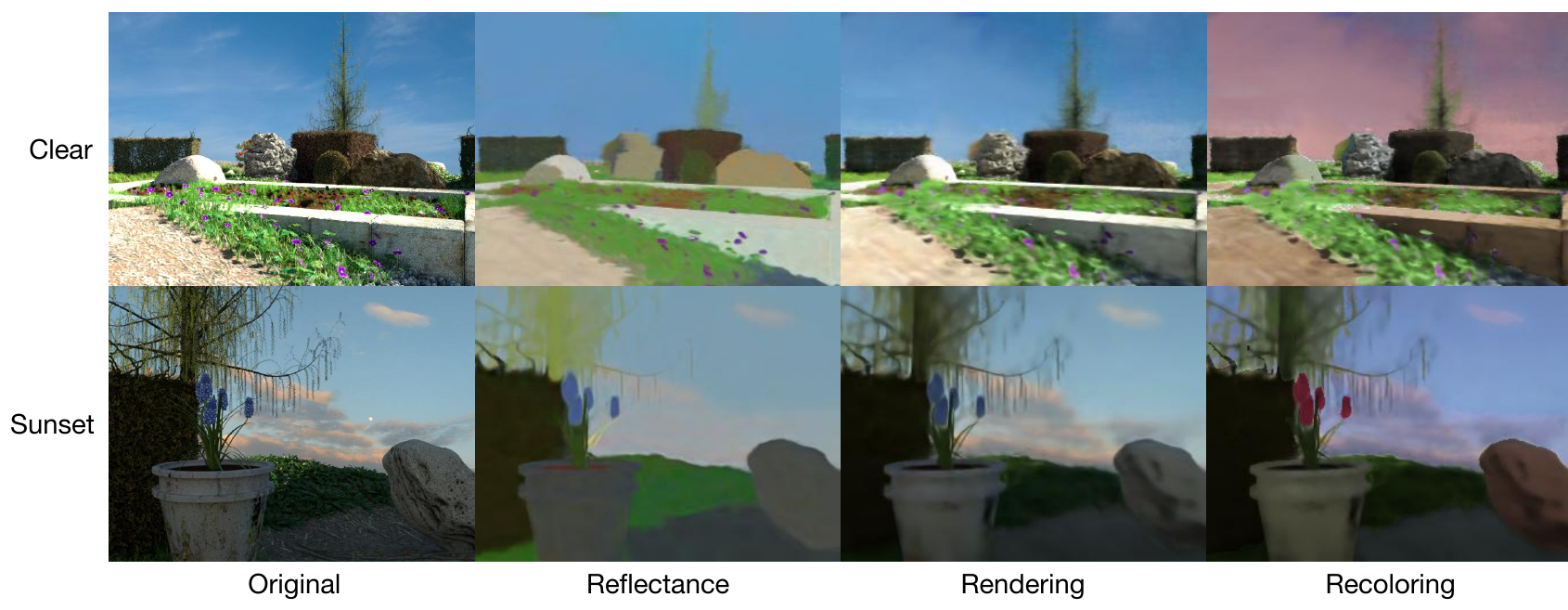}
\vspace{-1.8em}
\caption{Reflectance and Recoloring on the EDEN dataset.}
\label{fig:EDEN}
\vspace{-1.2em}
\end{figure}

\subsection{Applications}
\label{Sec:Applications} 
We demonstrate the applicability of IntrinsicNeRF with its decomposed components and the novel view synthesis results
on both synthetic and real-world data.



\begin{figure}[t!]
    \centering
    \includegraphics[width=1.0\linewidth, trim={0 0 0 0}, clip]{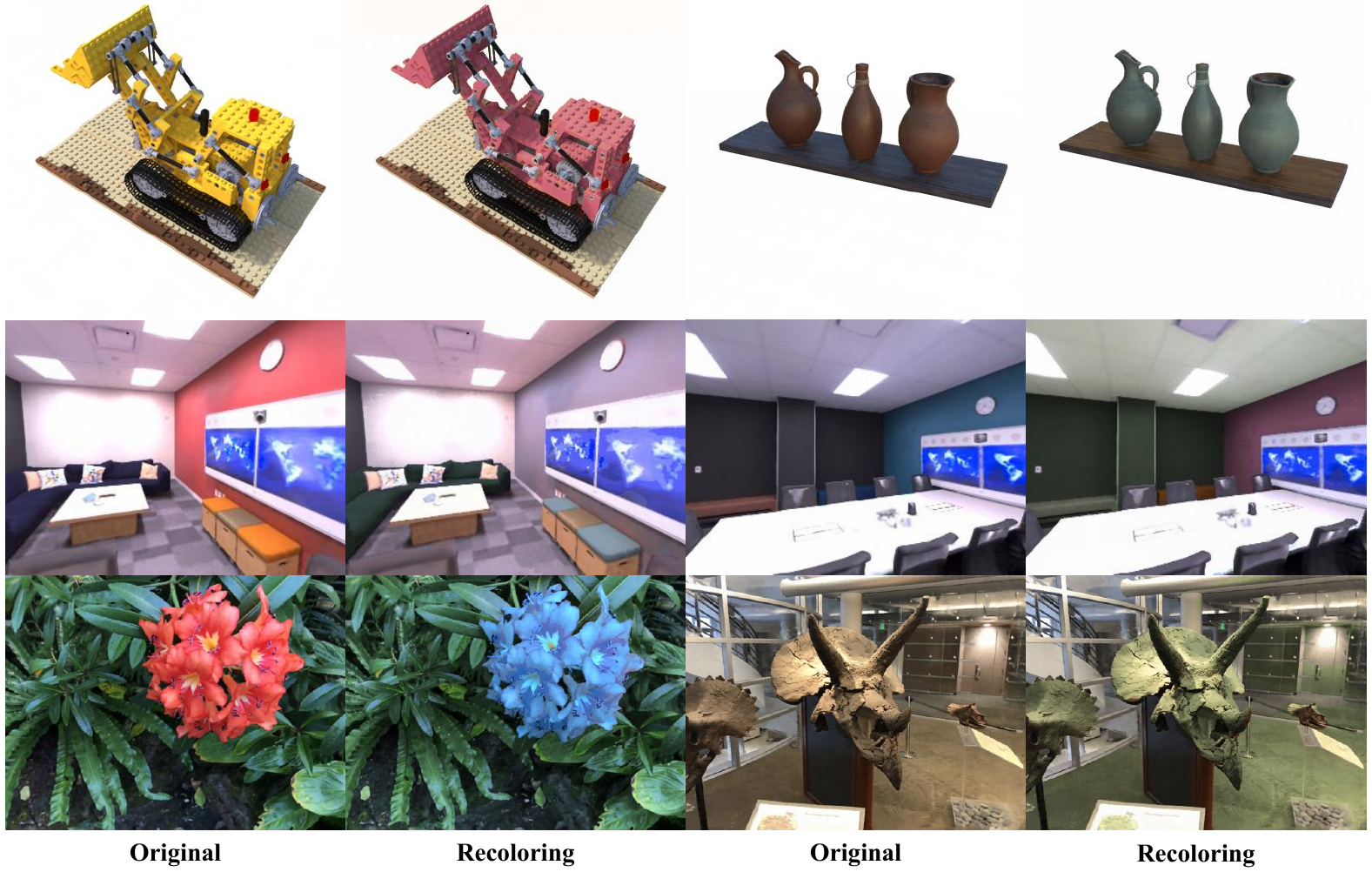}
    \vspace{-1.5em}
  \caption{\textbf{Recoloring on Synthetic/Real-World Data.}}
  \label{fig: Editing Examples on Synthetic and Real-World data}
  \vspace{-1em}
\end{figure}

\noindent\textbf{Scene Recoloring.}
\label{Sec:Real-Time Reflectance Recoloring}
In IntrinsicNeRF, the predicted reflectance is saved as [Semantic category, reflectance category] in the hierarchical iterative clustering and indexing method. We can simply modify the color of a certain reflectance category, the reflectance values of all pixels belonging to the selected category can be modified at the same time, and then the edited images can be reconstructed using the modified reflectance with the original shading and residual through Eq.~\ref{equa: intrinsic residual}. Fig.~\ref{fig: Editing Examples on Synthetic and Real-World data} shows some recoloring examples on both synthetic and real-world data. 

\begin{figure}[t!]
    \centering
    \includegraphics[width=1.0\linewidth, trim={0 0 0 0}, clip]{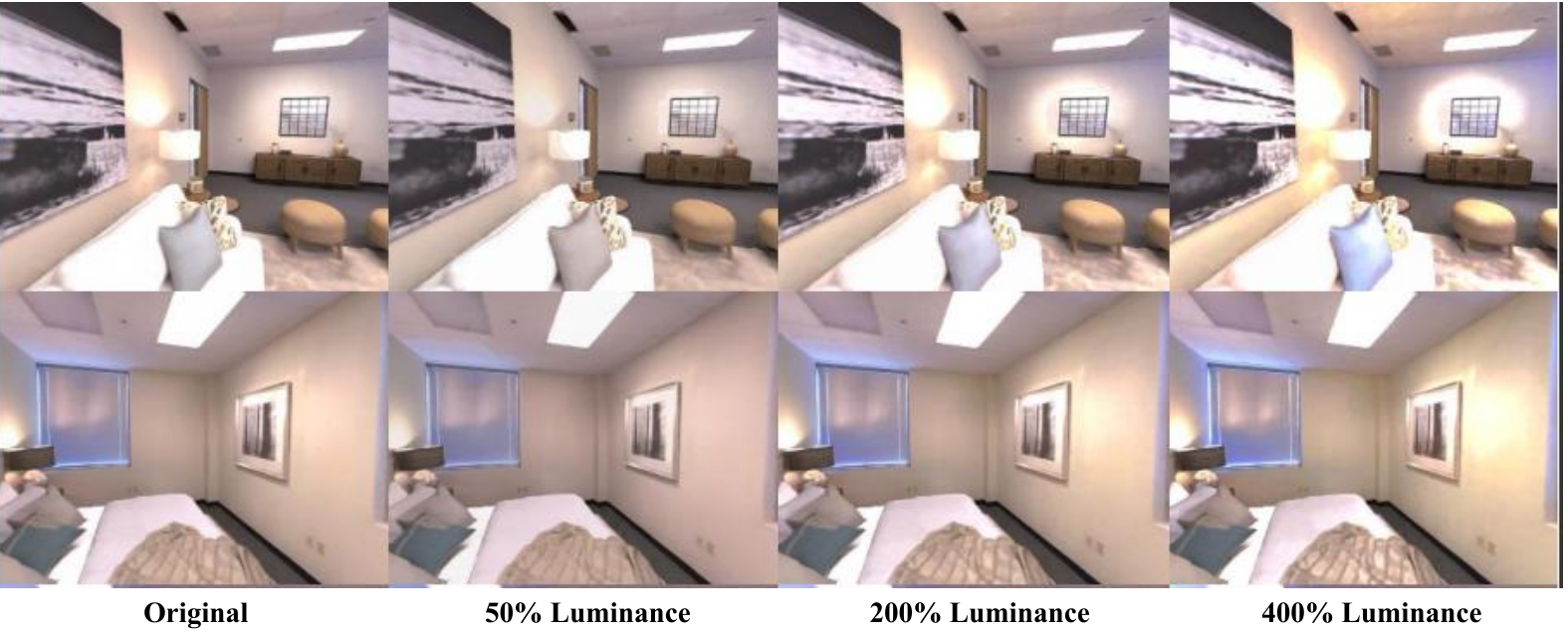}
    \vspace{-1.5em}
  \caption{\textbf{Illumination Variation on Replica Scene.}}
  \label{fig: Illumination Variation Samples}
  \vspace{-1em}
\end{figure}

\noindent\textbf{Illumination Variation.}
\label{Sec:Illumination Editing or Relighting}
The decomposed residual term beyond the Lambertian assumption, can represent the properties such as glossy illumination, we can adjust the overall brightness directly by a multiplicative factor. 
Fig.~\ref{fig: Illumination Variation Samples} shows the effect of different light intensities after enhancing or diminishing the light. Please see more edited samples and the novel view synthesis results 
in the supplementary material.
\section{Limitations and Future Work} The main limitation is that when the scenario does not conform to unsupervised intrinsic prior, it will struggle to obtain the correct decomposition results. A refinement method based on intrinsic decomposition prediction is required. Clustering errors may occur when the reflectance in a scene is complex and similar, especially in the real-world lacking semantic constraints. This can be solved by unsupervised semantic segmentation~\cite{hamilton2022unsupervised}. Estimating the reflectance requires a trade-off between preserving the texture and modeling the shadows correctly. Although our method performs well on room-scale scenes, such as Replica Scene, when applied to outdoor scenes with large scene sizes and fewer images, e.g. EDEN~\cite{le2021eden} under various lighting (clear and sunset), IntrinsicNeRF may lead to detail loss in rendering results (Fig.~\ref{fig:EDEN}). Meanwhile, the reflectance estimation may fall into local optimality due to limited observations, shown in Tab.~\ref{tab:EDEN} (Note that~\cite{zhang2021physg, zhang2022modeling} all fail). This can be addressed by combining our method with outdoor NeRF works~\cite{rudnev2022nerfosr, tancik2022block}. 
Although IntrinsicNeRF gives NeRF the ability to model the basic properties of scenes, it retains other shortcomings of NeRF. Given the high degree of integration of our approach with NeRF, NeRF extensions can be seamlessly incorporated into our IntrinsicNeRF, such as NeRF in the wild~\cite{chen2022hallucinated, mildenhall2020nerf, sun2022neuconw}, NeRF in dynamic environments~\cite{li2021neural, park2021nerfies, pumarola2020d,  Ye2022DeFlowSLAM}, fast NeRF~\cite{mueller2022instant, garbin2021fastnerf, chen2022tensorf, yu2021plenoctrees}, NeRF with generalization~\cite{chen2021mvsnerf, wang2021ibrnet, yu2021pixelnerf, kwon2021neural}, generative NeRF~\cite{schwarz2020graf, trevithick2021grf}, NeRF with panoptic segmentation~\cite{KunduCVPR2022PNF, Ye2022PVO}, NeRF-based SLAM~\cite{ming2022idf, sucar2021imap, zhu2022nice}, Geometry and Texture Editing with NeRF~\cite{bao2023sine, neumesh} etc, which will be helpful to the community. Another interesting direction is to unify intrinsic decomposition and inverse rendering to construct a hierarchical representation of the intrinsic properties of the scene. 
Since our approach yields multi-view consistent intrinsic decomposition results, IntrinsicNeRF can improve the performance of the intrinsic decomposition method by providing more datasets with pseudo-Ground-Truth labels for the intrinsic decomposition task. We leave this as future work.




\section{Conclusion}
We introduce intrinsic decomposition into neural rendering and propose intrinsic neural radiance fields that can decompose the images into reflectance, shading, and residual layers. Several techniques are proposed to make decomposition learning feasible and support online augmented applications such as recoloring, illumination variation, and editable novel view synthesis. We believe our method is the step toward the intrinsic decomposition (beyond Lambertian assumption) of more general scenes with neural rendering and will inspire follow-up work.

\noindent\textbf{Acknowledgments.}
The authors thank Yuanqing Zhang for providing us with the pre-trained model of InvRender~\cite{zhang2022modeling}, Jiarun Liu for reproducing the results of NeRFactor~\cite{zhang2021nerfactor} and PhySG~\cite{zhang2021physg}. We thank Hai Li and Jundan Luo for proofreading the paper. This work was partially supported by NSF of China (No. 61932003) and ZJU-SenseTime Joint Lab of 3D Vision. Weicai Ye was partially supported by China Scholarship Council (No. 202206320316).

{\small
\bibliographystyle{ieee_fullname}
\bibliography{IntrinsicNeRF_arxiv_new}
}

\clearpage
\appendix

\setcounter{page}{13}

\twocolumn[
    \centering
    \Large
    \textbf{IntrinsicNeRF: Learning Intrinsic Neural Radiance Fields for Editable Novel View Synthesis} \\
    \vspace{0.5em}Supplementary Document \\
    \vspace{1.0em}
] 

\setcounter{table}{0}
\setcounter{figure}{0}
\setcounter{equation}{0}
\renewcommand{\thetable}{\thesection\arabic{table}}
\renewcommand{\thefigure}{\thesection\arabic{figure}}
\renewcommand{\theequation}{\thesection\arabic{equation}}

In this supplementary document, we provide the framework of the semantic branch in IntrinsicNeRF (Sec.~\ref{sec: supp.Semantic Branch in IntrinsicNeRF}), 
and more experimental results (Sec.~\ref{sec.supp.More Experimental Results}) such as qualitative and quantitative results on the Blender Object dataset (Sec.~\ref{supp.Comparison Results on Blender Object}) and the Replica Scene dataset (Sec.~\ref{supp.Comparison Results on Replica Scene.}), and ablation studies (Sec.~\ref{supp.Ablation Study.}). We also present the applicability of our method on both synthetic and real-world data (Sec.~\ref{supp.sec.applications}).

\section{Semantic Branch in IntrinsicNeRF}
\label{sec: supp.Semantic Branch in IntrinsicNeRF}
\begin{figure}[h]
  \centering
  \includegraphics[width=\linewidth]{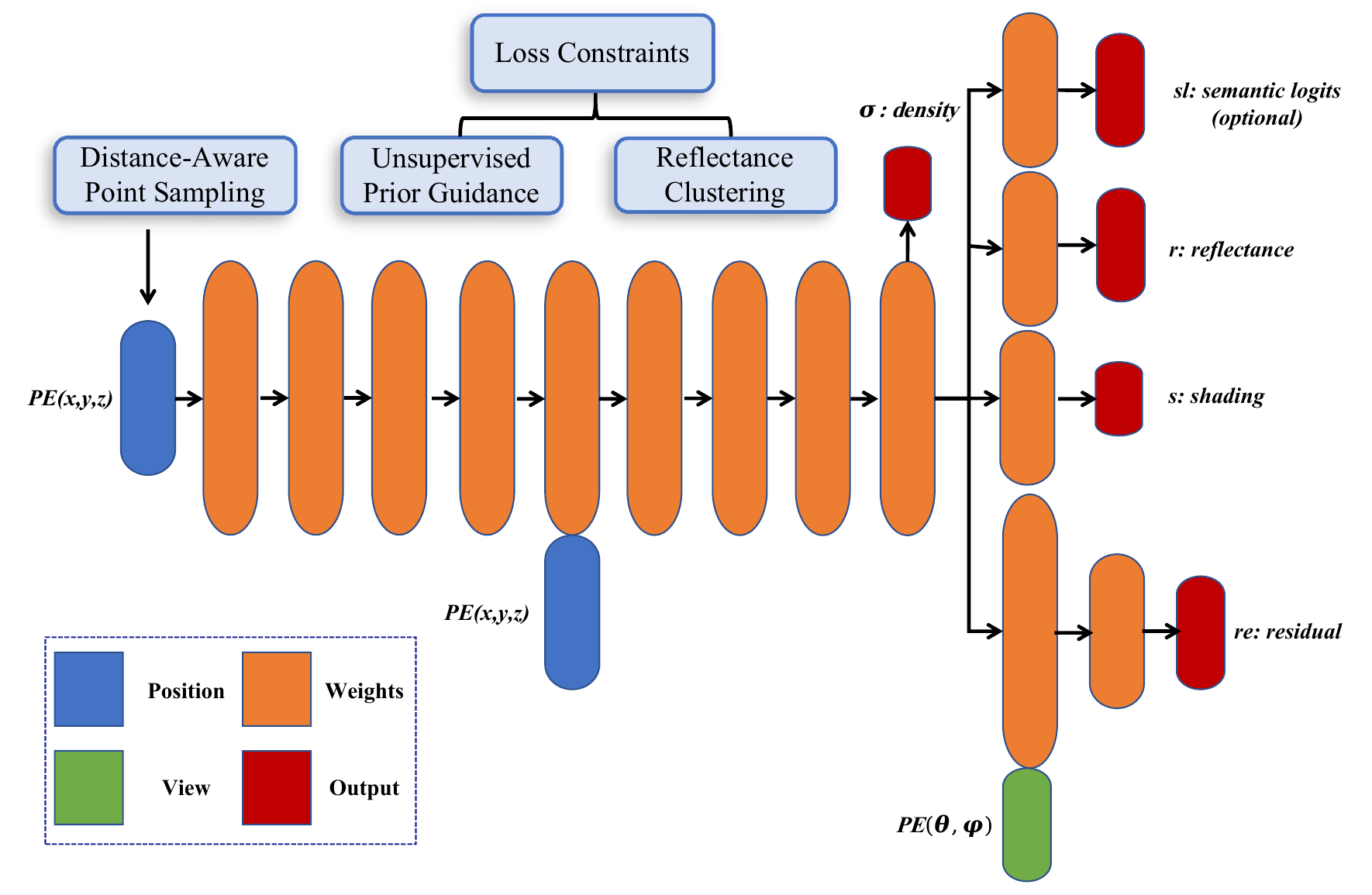}
  \caption{\textbf{IntrinsicNeRF Network.} IntrinsicNeRF takes 3D position $\mathbf{x}$=$(x,y,z)$ as input, and outputs view-independent volume density $\sigma$, semantic logits $sl$, reflectance $r$, and shading $s$. While the residual term $re$ additionally depends on the viewing direction $\mathbf{r}$=$(\theta, \phi)$. Distance-aware point sampling, unsupervised prior, and reflectance clustering methods are used to train the network.}
  \label{fig:supp.network details}
\end{figure}
Inspired by~\cite{zhi2021place}, we extend IntrinsicNeRF to jointly encode appearance, geometry, and semantics by appending a segmentation renderer to the original IntrinsicNeRF, shown in Fig.~\ref{fig:supp.network details}.
Following Semantic-NeRF~\cite{zhi2021place}, semantic segmentation is formalized as a view-independent function that recognized each pixel $\mathbf{x}$ as a semantic label distribution with softmax semantic logits $sl (\mathbf{x})$: 
\begin{equation} \label{eq:seg_renderer}
sl = F_{\Theta}(\mathbf{x}),
\end{equation}
where $F_{\Theta}$ is the MLP function. The predicted semantic logits $\hat{SL}(\mathbf{r})$ of each pixels can be written as:
\begin{align}
    \hat{{SL}}(\mathbf{r}) =&
        \sum_{k=1}^{K} \hat{T}_k \, \alpha_k \, sl_k \, \,
    \text{and}\, \hat{T}(t_k) = \exp \left(-\sum_{k'=1}^{k-1} \sigma_k \delta_k \right),
\end{align}
where $\alpha_k=1-\exp(-\sigma_k \delta_k)$, and $\delta_k$ is the distance between two adjacent sampled points along the view direction $\mathbf{r}$. Following Semantic-NeRF~\cite{zhi2021place}, we present semantic logits as multi-class probabilities with the cross-entropy loss:
\begin{align} 
L_{sem} & = -\sum_{\mathbf{r}\in \mathcal{R}} \left[
p\log \hat{p}_{c}+ p \log \hat{p}_{f}
\right],
\end{align}
where ${p}$ is the multi-class semantic probabilities of the ground truth semantic map, while $\hat{p}_{c}$ and $\hat{p}_{f}$ are the probabilities of coarse and fine predictions, respectively.

\section{More Experimental Results}
\label{sec.supp.More Experimental Results}
\begin{figure*}[h]
  \centering
  \includegraphics[width=\linewidth]{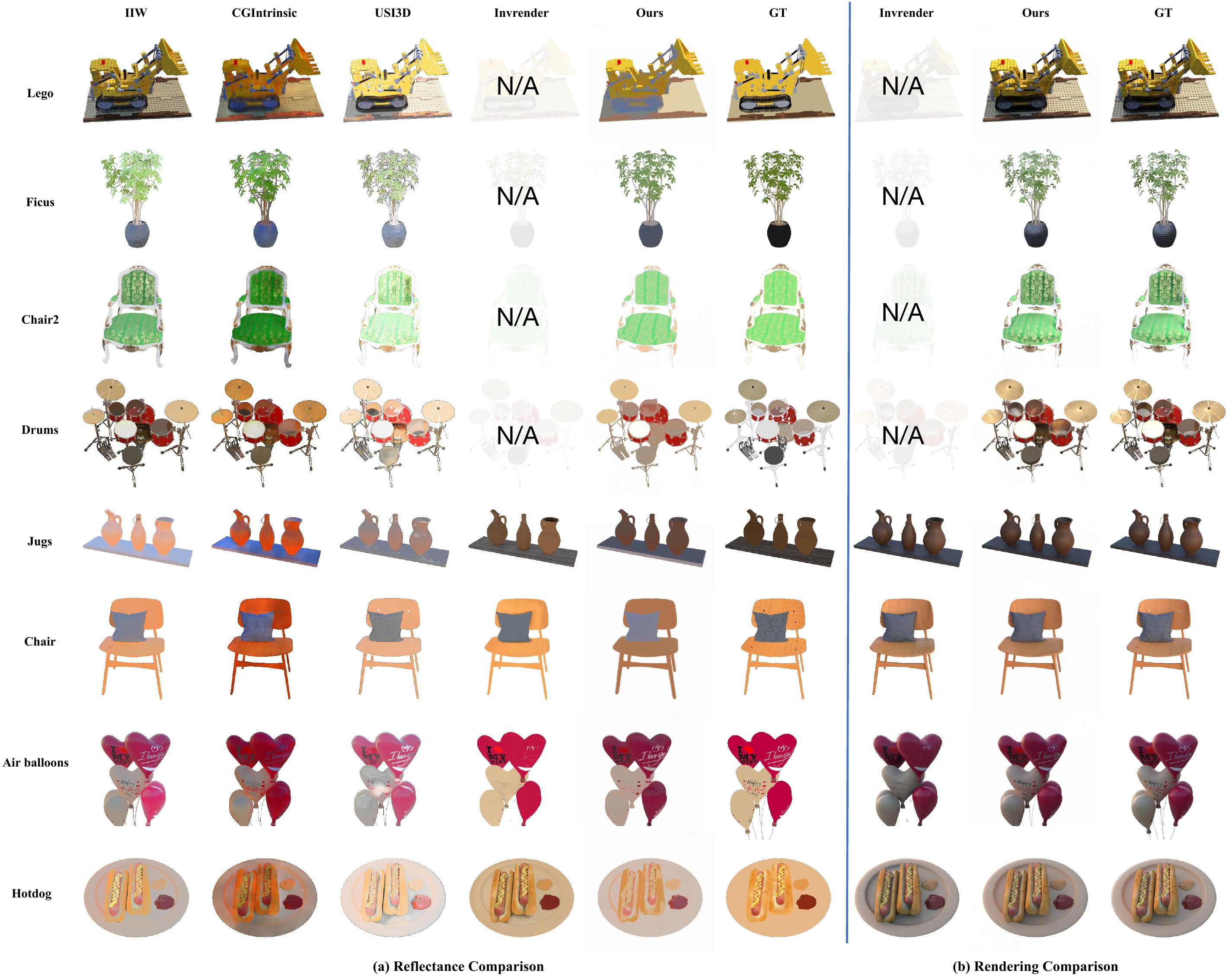}
  \vspace{-1.5em}
  \caption{\textbf{Qualitative Comparison Results of Reflectance and Rendering with Previous Work on the Blender Object Dataset.} The top 4 rows represent the sample of our dataset and the bottom 4 rows represent the sample of the Invrender dataset. Our method can perform reflectance estimation and novel view synthesis on both datasets well, while Invrender~\cite{zhang2022modeling} fails to do that on our dataset. N/A means failure.}
  \label{supp.fig: Comparisons with Previous work on Blender Datasets}
  \vspace{-1em}
\end{figure*}

\begin{figure}[h]
  \centering
  \includegraphics[width=\linewidth]{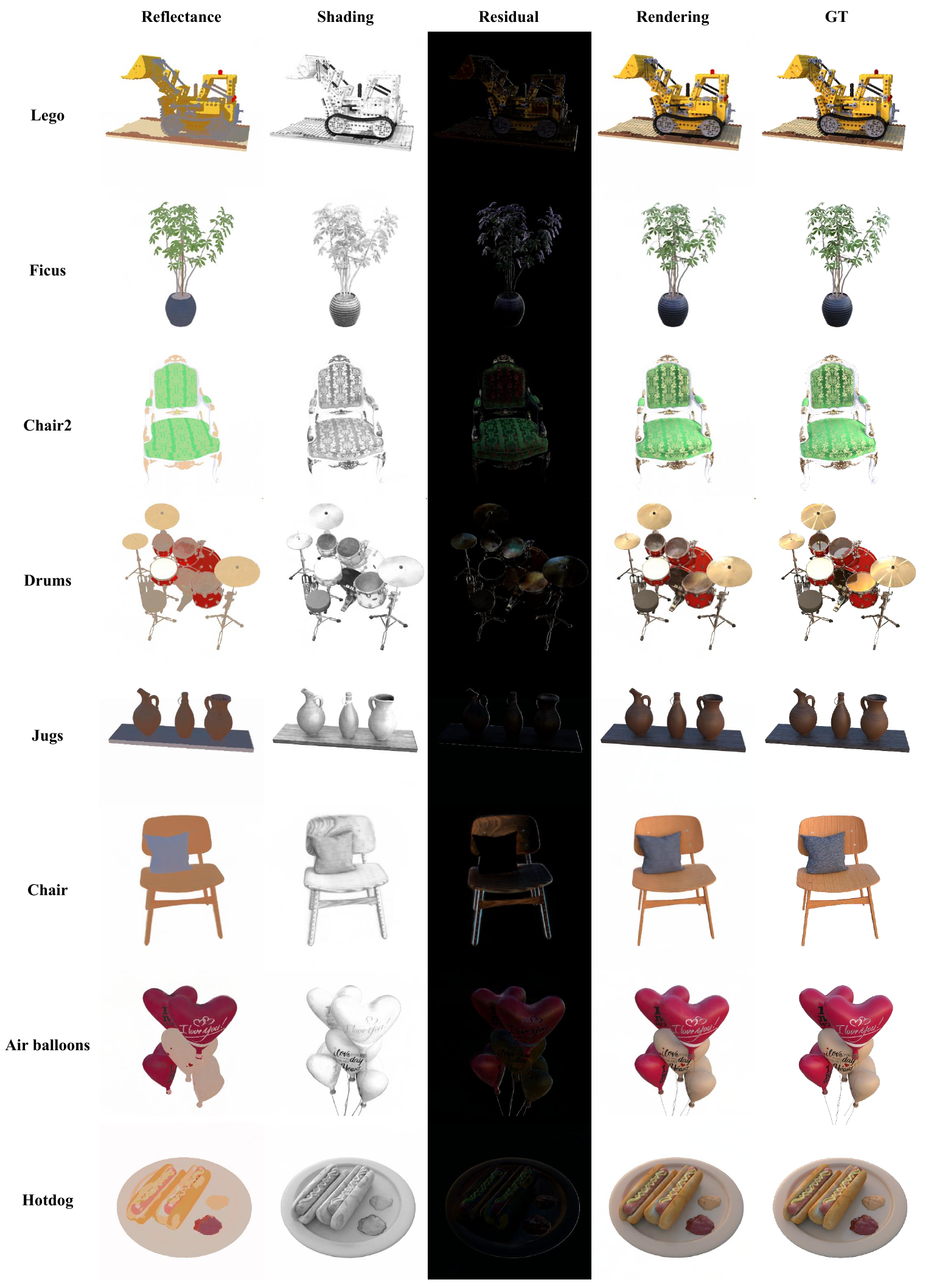}
  \vspace{-1.5em}
  \caption{\textbf{Qualitative Results of IntrinsicNeRF on the Blender Object Dataset.} From left to right are reflectance, shading, residual term, rendering result, and original image. In addition to the Lambertian assumption, our method can also simulate glossy reflections or metallic materials.}
  \label{supp.fig: Our intrinsic results on Blender Dataset}
\end{figure}

\begin{figure}[h]
  \centering
  \includegraphics[width=\linewidth]{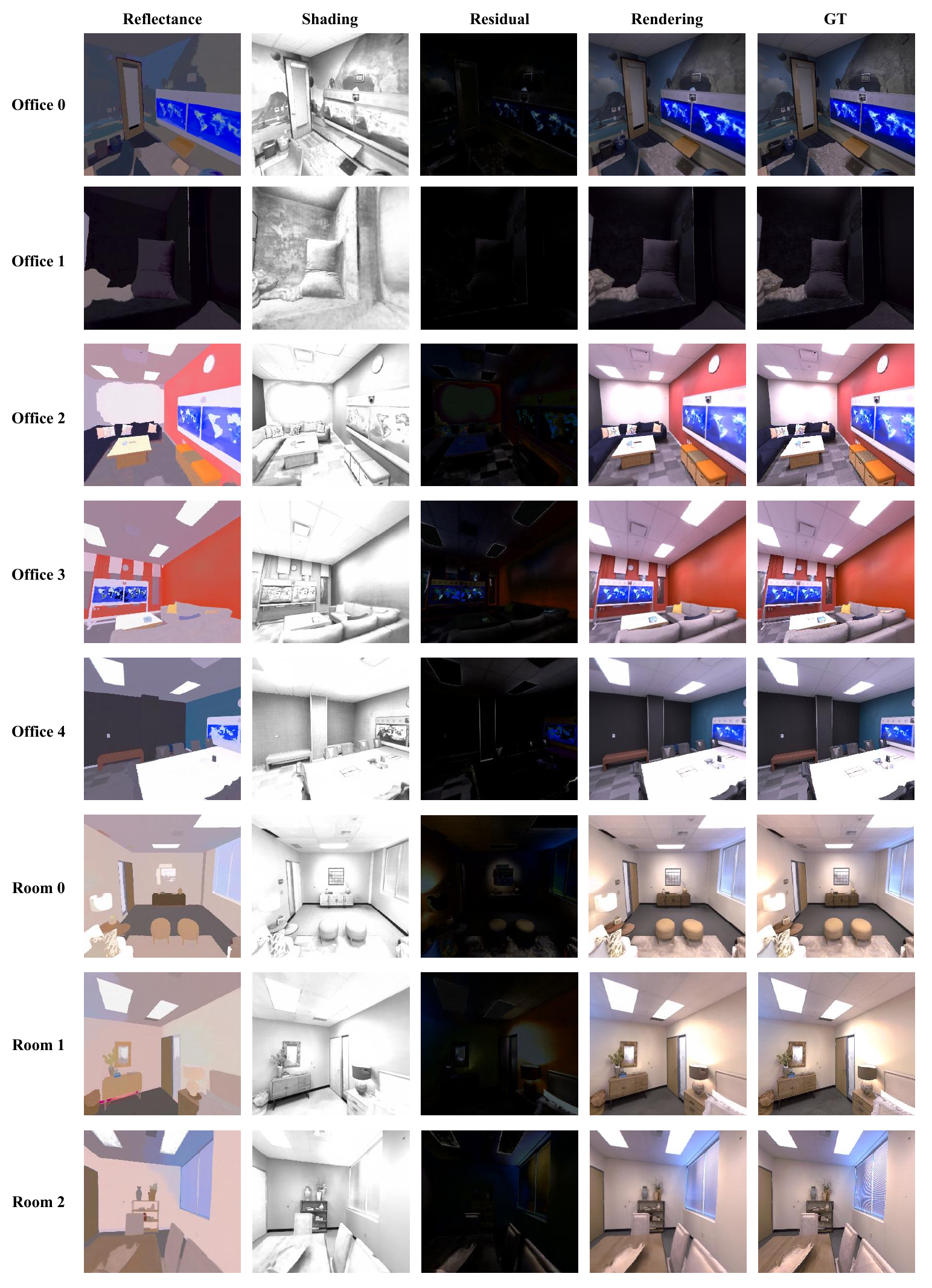}
  \vspace{-1.5em}
  \caption{\textbf{Qualitative Results of IntrinsicNeRF on the Replica Scene Dataset.} From left to right are reflectance, shading, residual term, rendering result, and original image. In addition to the Lambertian assumption, our method can also simulate glossy reflections or metallic materials.}
  \label{supp.fig: Our intrinsic results on Replica Scene}
\end{figure}

\begin{figure*}[!h] 
  \centering
  \includegraphics[width=\linewidth]{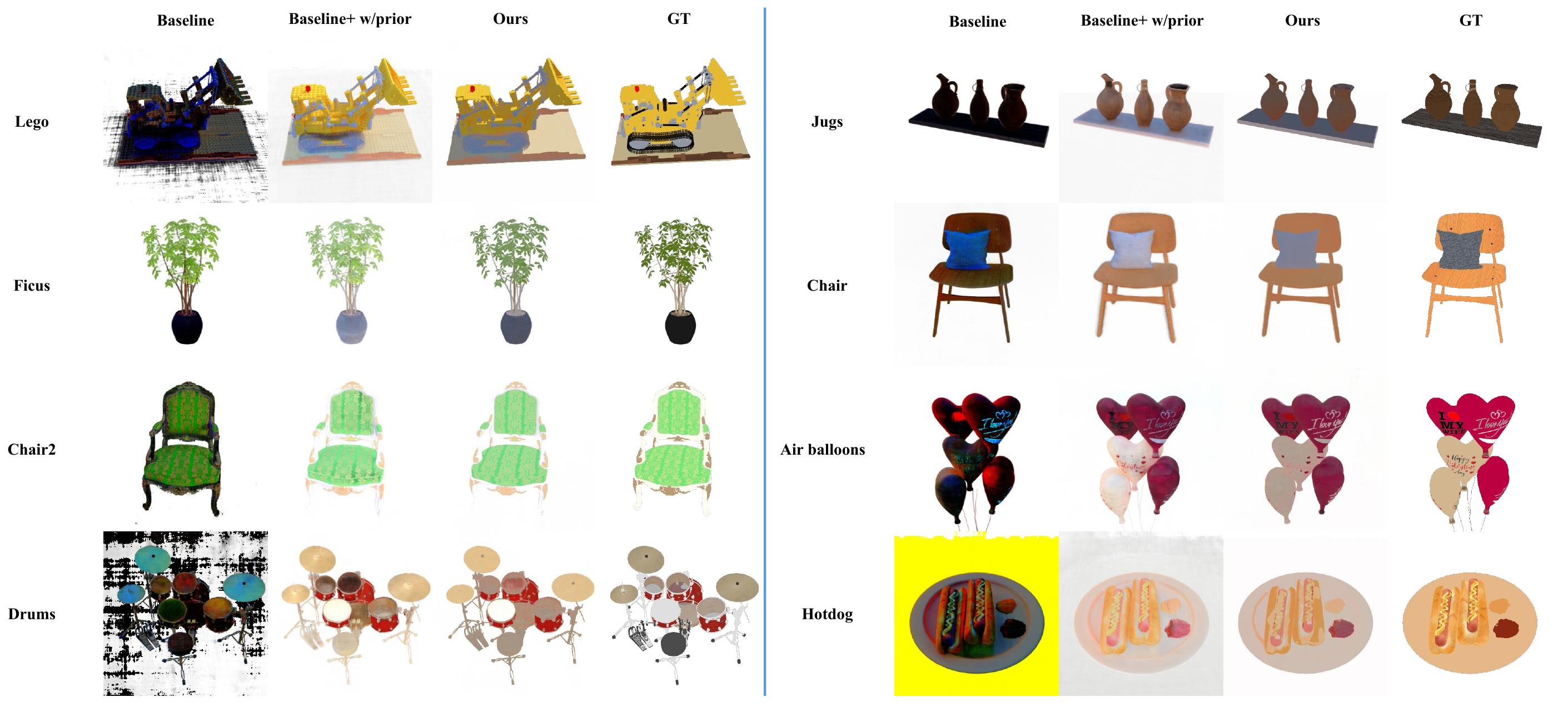}
  \caption{\textbf{Ablation study of Reflectance Estimation on the Blender Object Dataset.}    Left: our dataset, right: Invrender dataset. The reflectance estimation of the baseline method is stochastic and unstable, while the intrinsic prior makes the optimization of the network traceable. Our final model achieves more plausible reflectance results.}
  \label{supp.fig: Quality Results on Blender Dataset}
\end{figure*}

\begin{figure}[h]
  \centering
  \includegraphics[width=\linewidth]{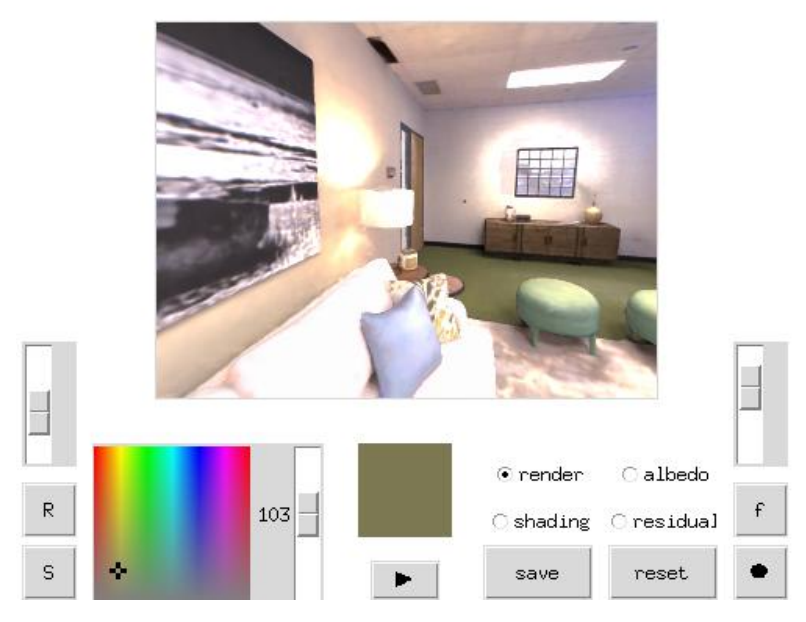}
  \caption{\textbf{Video Editing Software.} The software includes a palette for reflectance, a sliding bar for shading, residual layers, as well as buttons for playing or recording view synthesis, reset, etc.}
  \label{supp.fig: Video Editing Software}
\end{figure}

\begin{figure*}[h]
  \centering
\includegraphics[width=\linewidth]{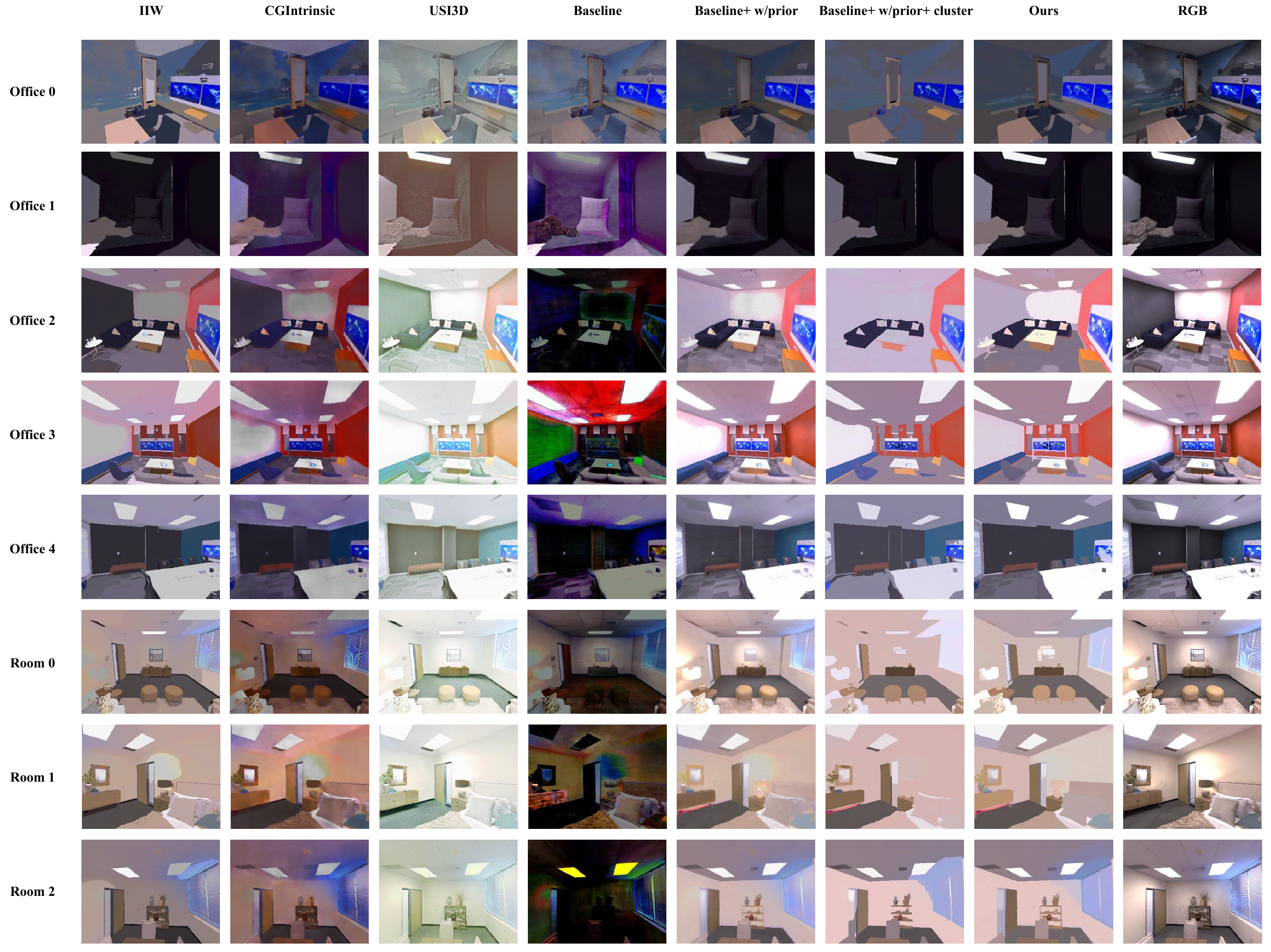}
  \caption{\textbf{Qualitative Reflectance Comparisons with Previous Methods on the Replica Scene Dataset.} Experiments demonstrate the progressive facilitation effect of our different variants. Compared with previous methods, our final method achieves more plausible and multi-view consistent reflectance estimation results, retaining the boundaries of objects, please refer to the supplementary video.}
  \label{supp.fig: Comparisons with Previous work on Replica Scene}
\end{figure*}

\subsection{Comparison on the Blender Object Dataset}
\label{supp.Comparison Results on Blender Object}
We present the detailed quantitative results on Tab.~\ref{tab: nerf dataset} and Tab.~\ref{tab: invrender dataset}, compared with intrinsic decomposition methods and neural rendering methods. Our full model is superior to existing traditional intrinsic decomposition methods such as USI3D~\cite{liu2020unsupervised}, IIW~\cite{bell2014intrinsic}, CGIntrinsic~\cite{li2018cgintrinsics} and reaches comparable results with Invrender~\cite{zhang2022modeling} in intrinsic decomposition on Invrender dataset, shown in Fig.~\ref{supp.fig: Comparisons with Previous work on Blender Datasets}. 
Furthermore, our intrinsic neural radiance field scene representation enhances reconstructing objects with complex shapes and textures on our dataset, while Invrender fails to make it. The qualitative results of IntrinsicNeRF on the Blender Object dataset are shown in Fig.~\ref{supp.fig: Our intrinsic results on Blender Dataset}. 
However, our method also falls into some local optima in Lego tracks (see Fig.~\ref{supp.fig: Quality Results on Blender Dataset}), due to the inherent property of the intrinsic decomposition, failing to handle the black regions. Meanwhile, when the scenario does not conform to unsupervised prior, it will struggle to obtain the correct decomposition results, as shown in Fig.~\ref{supp.fig: Comparisons with Previous work on Blender Datasets} (Hotdog, Chair in Ours column).

\begin{table*}[t]
\small
\begin{centering}
\scriptsize
\resizebox{\textwidth}{16mm}{
\begin{tabular}{ c c c c c c c c c c c c c c c c c c c}
\toprule
& \multicolumn{5}{c}{Reflectance (Lego)} & \multicolumn{3}{c}{View Synthesis (Lego)} &
 \multicolumn{5}{c}{Reflectance (Ficus)} & \multicolumn{3}{c}{View Synthesis (Ficus)}\\
 \cmidrule(lr){1-1}\cmidrule(lr){2-6}\cmidrule(lr){7-9}\cmidrule(lr){10-14} \cmidrule(lr){15-17}
 Method 
 & PSNR $\uparrow$ & SSIM $\uparrow$ & LPIPS $\downarrow$ & MSE $\downarrow$ & LMSE $\downarrow$ 
 & PSNR $\uparrow$ & SSIM $\uparrow$ & LPIPS $\downarrow$ 
 & PSNR $\uparrow$ & SSIM $\uparrow$ & LPIPS $\downarrow$ & MSE $\downarrow$ & LMSE $\downarrow$ 
 & PSNR $\uparrow$ & SSIM $\uparrow$ & LPIPS $\downarrow$ \\
\midrule
IIW~\cite{bell2014intrinsic} &\underline{21.3080} &0.8840 &\underline{0.1255} &\underline{0.0075} &0.0355 &- &- &-  &19.4159 &0.9145 &0.0803 &0.0110 &0.1330 &- &- &- \\
CGIntrinsic~\cite{li2018cgintrinsics} &18.6028 &0.8683 &0.1454 &0.0123 &0.0363 &- &- &- &22.0665 &\textbf{0.9408} &0.0513 &0.0052 &0.1298 &- &- &-\\
USI3D~\cite{liu2020unsupervised} &18.2291 &0.8822 &0.1282 &0.0146 &\underline{0.0332} &- &- &- &16.2838 &0.9253 &0.0746 &0.0230 &0.0995 &- &- &- \\


\midrule
NeRFactor~\cite{zhang2021nerfactor} & \textbf{22.5591}	&\textbf{0.9250} &\textbf{0.0875}	&\textbf{0.0034}	&\textbf{0.0262} &17.6665	&0.8263	&0.1504 & 19.6809	&0.9107	&0.0488	&0.0104	&\underline{0.0874} &21.3010	&0.9053	&0.0678 \\
PhySG~\cite{zhang2021physg} &- &- &- &- &- &- &- &- &- &- &- &- &- &- &- &- \\
Invrender~\cite{zhang2022modeling} &- &- &- &- &- &- &- &- &- &- &- &- &- &- &- &- \\
NeRF~\cite{mildenhall2020nerf} &- &- &- &- &- &\textbf{29.5691} &\textbf{0.9331} &\textbf{0.0268} &- &- &- &- &- &\textbf{29.4080} &\textbf{0.9609} &\textbf{0.0155} \\
\midrule
baseline &11.9473 &0.7669 &0.2399 &0.0522 &0.2398 &\underline{29.4163} &\underline{0.9326} &\underline{0.0280} &\underline{23.0957} &0.9229 &\underline{0.0420} &\underline{0.0045} &0.1158 &\underline{29.3302} &\underline{0.9597} &\underline{0.0158} \\
baseline+w/prior &18.3652 &0.8832 &0.1515 &0.0136 &0.0615 &29.1918 &0.9300 &0.0313 &19.3838 &0.9232 &0.0606 &0.0112 &0.0933 &29.0722 &0.9588 &0.0170 \\
Ours &19.0001 &\underline{0.9046} &0.1288 &0.0116 &0.0647 &29.1526 &0.9283 &0.0308 &\textbf{23.3383} &\underline{0.9402} &\textbf{0.0325} &\textbf{0.0042} &\textbf{0.0676} &28.9046 &0.9576 &0.0175 \\
\midrule
\end{tabular}}

\scriptsize
\resizebox{\textwidth}{16mm}{
\begin{tabular}{ c c c c c c c c c c c c c c c c c c c}
\toprule
& \multicolumn{5}{c}{Reflectance (Chair2)} & \multicolumn{3}{c}{View Synthesis (Chair2)} &
 \multicolumn{5}{c}{Reflectance (Drums)} & \multicolumn{3}{c}{View Synthesis (Drums)}\\
 \cmidrule(lr){1-1}\cmidrule(lr){2-6}\cmidrule(lr){7-9}\cmidrule(lr){10-14} \cmidrule(lr){15-17}
 Method 
 & PSNR $\uparrow$ & SSIM $\uparrow$ & LPIPS $\downarrow$ & MSE $\downarrow$ & LMSE $\downarrow$ 
 & PSNR $\uparrow$ & SSIM $\uparrow$ & LPIPS $\downarrow$ 
 & PSNR $\uparrow$ & SSIM $\uparrow$ & LPIPS $\downarrow$ & MSE $\downarrow$ & LMSE $\downarrow$ 
 & PSNR $\uparrow$ & SSIM $\uparrow$ & LPIPS $\downarrow$ \\
\midrule
IIW~\cite{bell2014intrinsic} &24.2352 &\underline{0.9410} &0.0913 &0.0035 &0.0133 &- &- &- &17.1604 &0.8918 &\underline{0.1553} &0.0188 &0.1091 &- &- &- \\
CGIntrinsic~\cite{li2018cgintrinsics} &15.9210 &0.9070 &0.1363 &0.0259 &0.0265 &- &- &- &17.1604 &0.8918 &\underline{0.1553} &0.0188 &0.1091 &- &- &- \\
USI3D~\cite{liu2020unsupervised} &23.0661 &0.9303 &0.1092 &0.0045 &0.0108 &- &- &- &16.8267 &0.8835 &0.1588 &0.0188 &0.0711 &- &- &- \\


\midrule
NeRFactor~\cite{zhang2021nerfactor} &21.5867	&0.9266	&0.1680	&0.0056	&0.0203 & 25.5135	&0.8919	&0.1285 &\textbf{21.9491}	&0.9059	&\textbf{0.1176}	&\textbf{0.0059}	&\textbf{0.0438} &20.6880 	&0.8733	&0.1185 \\
PhySG~\cite{zhang2021physg} &- &- &- &- &- &- &- &- &- &- &- &- &- &- &- &- \\
Invrender~\cite{zhang2022modeling} &- &- &- &- &- &- &- &- &- &- &- &- &- &- &- &- \\
NeRF~\cite{mildenhall2020nerf} &- &- &- &- &- &\textbf{30.1428} &\textbf{0.9448} &\textbf{0.0301} &- &- &- &- &- &\textbf{24.4357} &\textbf{0.9205} &\textbf{0.0590} \\
\midrule
baseline &11.0799 &0.8387 &0.2025 &0.0810 &0.1802 &\underline{30.0731} &\underline{0.9436} &\underline{0.0304} &13.3059 &0.8301 &0.2110 &0.0426 &0.2036 &\underline{24.2220} &\underline{0.9172} &\underline{0.0614} \\
baseline+w/prior &\underline{27.1114} &0.9406 &\underline{0.0897} &\underline{0.0015} &\underline{0.0067} &29.7973 &0.9406 &0.0368 &18.9980 &\underline{0.9089} &0.1845 &0.0117 &0.0537 &24.1918 &0.9188 &0.0625 \\
Ours &\textbf{28.0020} &\textbf{0.9486} &\textbf{0.0731} &\textbf{0.0011} &\textbf{0.0054} &29.6453 &0.9388 &0.0383 &\underline{19.9305} &\textbf{0.9133} &0.1555 &\underline{0.0093} &\underline{0.0518} &24.0949 &0.9182 &0.0620 \\
\midrule
\end{tabular}}
\caption{\textbf{Quantitative Evaluations on Our dataset.} Bold indicates best and underline indicates second best. - means failure.}\label{tab: nerf dataset}
\end{centering}
\end{table*}

\begin{table*}[t]
\small
\begin{centering}
\scriptsize
\resizebox{\textwidth}{16mm}{
\begin{tabular}{ c c c c c c c c c c c c c c c c c c c}
\toprule
& \multicolumn{5}{c}{Reflectance (Jugs)} & \multicolumn{3}{c}{View Synthesis (Jugs)} &
 \multicolumn{5}{c}{Reflectance (Chair)} & \multicolumn{3}{c}{View Synthesis (Chair)}\\
 \cmidrule(lr){1-1}\cmidrule(lr){2-6}\cmidrule(lr){7-9}\cmidrule(lr){10-14} \cmidrule(lr){15-17}
 Method 
 & PSNR $\uparrow$ & SSIM $\uparrow$ & LPIPS $\downarrow$ & MSE $\downarrow$ & LMSE $\downarrow$ 
 & PSNR $\uparrow$ & SSIM $\uparrow$ & LPIPS $\downarrow$ 
 & PSNR $\uparrow$ & SSIM $\uparrow$ & LPIPS $\downarrow$ & MSE $\downarrow$ & LMSE $\downarrow$ 
 & PSNR $\uparrow$ & SSIM $\uparrow$ & LPIPS $\downarrow$ \\
\midrule
IIW~\cite{bell2014intrinsic} &15.2941 &0.9105 &0.1188 &0.0320 &0.0238 &- &- &-  &\underline{25.8220} &\textbf{0.9337} &\underline{0.0620} &0.0019 &0.0091 &- &- &- \\
CGIntrinsic~\cite{li2018cgintrinsics} &19.2596 &0.9313 &0.1066 &0.0086 &\underline{0.0220} &- &- &-&21.1657 &0.9140 &0.0855 &0.0070 &0.0098 &- &- &- \\
USI3D~\cite{liu2020unsupervised} &18.4617 &0.9242 &0.0780 &0.0147 &0.0249 &- &- &- &24.5503 &\underline{0.9290} &0.0744 &0.0020 &\textbf{0.0070} &- &- &- \\


\midrule
NeRFactor~\cite{zhang2021nerfactor} &19.1639	&0.9275	&0.0911	&0.0116	&\textbf{0.0215} &26.0967	&0.9492	&0.0430 & 22.0620	&0.9208	&0.1287	&\underline{0.0014}	&0.0089 & 22.1625	&0.9294	&0.0876 \\
PhySG~\cite{zhang2021physg} & 24.6498 & 0.9427 & 0.0790 & 0.0034 & 0.0860 & 24.6221 & 0.9544 & 0.0609 & 24.9832 & 0.9168 & 0.0877 & 0.0024 & 0.0262 &25.7197 & 0.9320 &0.0710 \\
Invrender~\cite{zhang2022modeling} &\underline{24.8413} &\textbf{0.9508} &\textbf{0.0361} &\underline{0.0033} &0.0427 &29.5990 &0.9654 &0.0266 &\textbf{29.4776} &0.9285 &\textbf{0.0574} &\textbf{0.0010} &\underline{0.0089} &31.3660 &0.9444 &0.0464 \\
NeRF~\cite{mildenhall2020nerf} &- &- &- &- &- &\textbf{35.4846} &\underline{0.9796} &\underline{0.0165} &- &- &- &- &- &32.5685 &0.9436 &0.0427 \\
\midrule
baseline &21.6691 &0.8750 &0.0773 &0.0065 &0.4158 &\underline{35.2488} &\textbf{0.9800} &\textbf{0.0155} &14.8468 &0.8679 &0.1271 &0.0277 &0.1151 &\textbf{34.1195} &\textbf{0.9522} &\textbf{0.0312} \\
baseline+w/prior &19.1960 &0.9249 &0.1136 &0.0117 &0.0331 &35.0930 &0.9769 &0.0212 &22.5096 &0.9232 &0.0875 &0.0042 &0.0156 &\underline{32.7608} &\underline{0.9445} &0.0424 \\
Ours &\textbf{25.7546} &\underline{0.9471} &\underline{0.0661} &\textbf{0.0025} &0.0308 &35.0342 &0.9769 &0.0213 &23.7306 &0.9278 &0.0854 &0.0027 &0.0110 &32.6955 &0.9441 &\underline{0.0415} \\
\midrule
\end{tabular}}

\scriptsize
\resizebox{\textwidth}{16mm}{
\begin{tabular}{ c c c c c c c c c c c c c c c c c c c}
\toprule
& \multicolumn{5}{c}{Reflectance (Air balloons)} & \multicolumn{3}{c}{View Synthesis (Air balloons)} &
 \multicolumn{5}{c}{Reflectance (Hotdog)} & \multicolumn{3}{c}{View Synthesis (Hotdog)}\\
 \cmidrule(lr){1-1}\cmidrule(lr){2-6}\cmidrule(lr){7-9}\cmidrule(lr){10-14} \cmidrule(lr){15-17}
 Method 
 & PSNR $\uparrow$ & SSIM $\uparrow$ & LPIPS $\downarrow$ & MSE $\downarrow$ & LMSE $\downarrow$ 
 & PSNR $\uparrow$ & SSIM $\uparrow$ & LPIPS $\downarrow$ 
 & PSNR $\uparrow$ & SSIM $\uparrow$ & LPIPS $\downarrow$ & MSE $\downarrow$ & LMSE $\downarrow$ 
 & PSNR $\uparrow$ & SSIM $\uparrow$ & LPIPS $\downarrow$ \\
\midrule
IIW~\cite{bell2014intrinsic} &22.4801 &\textbf{0.9276} &\textbf{0.0571} &0.0040 &\textbf{0.0087} &- &- &- &24.5176 &0.9512 &0.1009 &0.0014 &0.0062 &- &- &- \\
CGIntrinsic~\cite{li2018cgintrinsics} &20.6844 &0.9083 &0.0888 &0.0066 &0.0192 &- &- &- &19.5237 &0.9299 &0.1176 &0.0294 &0.0054 &- &- &- \\
USI3D~\cite{liu2020unsupervised} &19.2599 &0.9119 &0.0725 &0.0088 &\underline{0.0185} &- &- &- &20.7564 &0.9418 &0.1297 &0.0061 &0.0084 &- &- &- \\


\midrule
NeRFactor~\cite{zhang2021nerfactor} & 17.5734	&0.8770	&0.1701	&0.0063	&0.0416 & 20.7204	&0.9018	&0.1096 &20.8677 &0.9372	&0.1517	&0.0044	&0.0121  &23.0737	&0.9305	&0.0885 \\
PhySG~\cite{zhang2021physg} & \underline{22.7754} & 0.9080 & 0.0974 & 0.0035 & 0.0328 & 26.1276 & 0.9475 & 0.0781 & 21.0910 & 0.9248 & 0.1729 & 0.0042 & 0.0134 & 25.2207 & 0.9213 & 0.1115 \\
Invrender~\cite{zhang2022modeling} &\textbf{25.2053} &\underline{0.9155} &\underline{0.0716} &\underline{0.0026} &0.0263 &27.6636 &0.9493 &0.0779 &\textbf{25.7069} &0.9570 &\textbf{0.0637} &\underline{0.0020} &0.0123 &28.9192 &0.9497 &0.0513 \\
NeRF~\cite{mildenhall2020nerf} &- &- &- &- &- &\textbf{32.8084} &\textbf{0.9676} &\textbf{0.0224} &- &- &- &- &- &\textbf{34.2531} &\textbf{0.9697} &\textbf{0.0287} \\
\midrule
baseline &15.2960 &0.8601 &0.1399 &0.0241 &0.1820 &\underline{32.5626} &\underline{0.9666} &\underline{0.0251} &13.4718 &0.8517 &0.1762 &0.0432 &0.0690 &\underline{34.0833} &\underline{0.9693} &\underline{0.0292} \\
baseline+w/prior &21.2049 &0.9049 &0.1148 &0.0036 &0.0214 &32.3400 &0.9661 &0.0254 &24.0375 &\underline{0.9581} &0.1184 &0.0024 &\underline{0.0042} &33.7700 &0.9678 &0.0325 \\
Ours &21.9558 &0.9116 &0.1036 &\textbf{0.0023} &0.0235 &32.2197 &0.9648 &0.0269 &\underline{25.6160} &\textbf{0.9620} &\underline{0.0967} &\textbf{0.0008} &\textbf{0.0038} &34.0375 &0.9662 &0.0325 \\
\midrule
\end{tabular}}
\caption{\textbf{Quantitative Evaluations on Invrender dataset.} Bold indicates best and underline indicates second best. - means failure.}\label{tab: invrender dataset}
\vspace{-1.5em}
\end{centering}
\end{table*}
\begin{table*}[t]
\small
\begin{centering}

\scriptsize
\resizebox{\textwidth}{8mm}{
\begin{tabular}{ c c c c c c c c c c c c c c c c c c c}
\toprule
& \multicolumn{4}{c}{Office 0} & \multicolumn{4}{c}{Office 1} &
 \multicolumn{4}{c}{Office 2} & \multicolumn{4}{c}{Office 3}\\
 \cmidrule(lr){1-1}\cmidrule(lr){2-5}\cmidrule(lr){6-9}\cmidrule(lr){10-13} \cmidrule(lr){14-17}
 Method 
 & PSNR $\uparrow$ & SSIM $\uparrow$ & LPIPS $\downarrow$ & mIoU$\uparrow$ 
 & PSNR $\uparrow$ & SSIM $\uparrow$ & LPIPS $\downarrow$ & mIoU$\uparrow$ 
 & PSNR $\uparrow$ & SSIM $\uparrow$ & LPIPS $\downarrow$ & mIoU$\uparrow$ 
 & PSNR $\uparrow$ & SSIM $\uparrow$ & LPIPS $\downarrow$ & mIoU$\uparrow$  \\
\midrule

Semantic-NeRF~\cite{zhi2021place} 
&33.9807 &0.9294 &0.0631 &0.9802
&35.6869 &0.9516 &0.0689 &0.9816
&30.8175 &0.9296 &0.0755 &0.9777
&30.2418 &0.9238 &0.0694 &0.9678 \\

Ours 
&33.9734 &0.9292 &0.0666 &0.9793
&35.4500 &0.9532 &0.0680 &0.9809
&30.2827 &0.9231 &0.0843 &0.9753
&29.9553 &0.9179 &0.0741 &0.9619 \\

\midrule
\end{tabular}}

\scriptsize
\resizebox{\textwidth}{8mm}{
\begin{tabular}{ c c c c c c c c c c c c c c c c c c c}
\toprule
& \multicolumn{4}{c}{Office 3} & \multicolumn{4}{c}{Room 0} &
 \multicolumn{4}{c}{Room 1} & \multicolumn{4}{c}{Room 2}\\
 \cmidrule(lr){1-1}\cmidrule(lr){2-5}\cmidrule(lr){6-9}\cmidrule(lr){10-13} \cmidrule(lr){14-17}
 Method 
 & PSNR $\uparrow$ & SSIM $\uparrow$ & LPIPS $\downarrow$ & mIoU$\uparrow$ 
 & PSNR $\uparrow$ & SSIM $\uparrow$ & LPIPS $\downarrow$ & mIoU$\uparrow$ 
 & PSNR $\uparrow$ & SSIM $\uparrow$ & LPIPS $\downarrow$ & mIoU$\uparrow$ 
 & PSNR $\uparrow$ & SSIM $\uparrow$ & LPIPS $\downarrow$ & mIoU$\uparrow$  \\
\midrule

Semantic-NeRF~\cite{zhi2021place} 
&31.4142 &0.9154 &0.1039 &0.9531
&27.2094 &0.8108 &0.1669 &0.9712
&28.5790 &0.8215 &0.1719 &0.9802
&29.8863 &0.8814 &0.1331 &0.9681

\\
Ours 
&30.9201 &0.9106 &0.1098 &0.9537
&27.0812 &0.8063 &0.1698 &0.9680
&28.1852 &0.8048 &0.2056 &0.9769
&29.7873 &0.8809 &0.1343 &0.9651

\\
\midrule
\end{tabular}}
\caption{\textbf{Quantitative Evaluations on the Replica Scene Dataset.}  We achieve comparable results with Semantic-NeRF in novel view synthesis and semantic segmentation.}\label{tab: replica with semantic-nerf}
\end{centering}
\end{table*}

\subsection{Comparison on the Replica Scene Dataset}
\label{supp.Comparison Results on Replica Scene.}
Tab.~\ref{tab: replica with semantic-nerf} shows the complete quantitative results on the Replica Scene dataset for novel view synthesis and semantic segmentation. We achieve comparable results with Semantic-NeRF~\cite{zhi2021place} while giving the ability to model the underlying properties of scenes. Fig.~\ref{supp.fig: Our intrinsic results on Replica Scene} shows the qualitative results of IntrinsicNeRF on the Replica Scene dataset.

\subsection{Ablation Studies}
\label{supp.Ablation Study.}
We show more ablation study results in Fig.~\ref{supp.fig: Quality Results on Blender Dataset} on the Blender Object dataset and in Fig.~\ref{supp.fig: Comparisons with Previous work on Replica Scene} on the Replica Scene dataset. The reflectance estimated by the baseline method is more stochastic and unstable. While adding the intrinsic prior, the network output is plausible. The adaptive reflectance iterative clustering method can make the reflectance regions of the same material cluster together but may lose some distinguishable boundaries in the Replica Scene dataset. We also show the quantitative comparison results of the Blender Object dataset in Tab.~\ref{tab: nerf dataset} and Tab.~\ref{tab: invrender dataset}. The comparison results demonstrate that unsupervised prior and clustering can help to improve the intrinsic decomposition, but may decrease the performance of view synthesis slightly. Fig.~\ref{supp.fig: Comparisons with Previous work on Replica Scene} shows hierarchical clustering method can retain the boundaries and still yields more plausible results. 

\subsection{Applications}
\label{supp.sec.applications}
We show the applicability of IntrinsicNeRF on real-time scene recoloring, illumination variation, and editable novel view synthesis. We have also developed a convenient editing software, to facilitate the user to perform object or scene editing, shown in Fig.~\ref{supp.fig: Video Editing Software}.

\begin{figure*}[h]
  \centering
  \includegraphics[width=\linewidth]{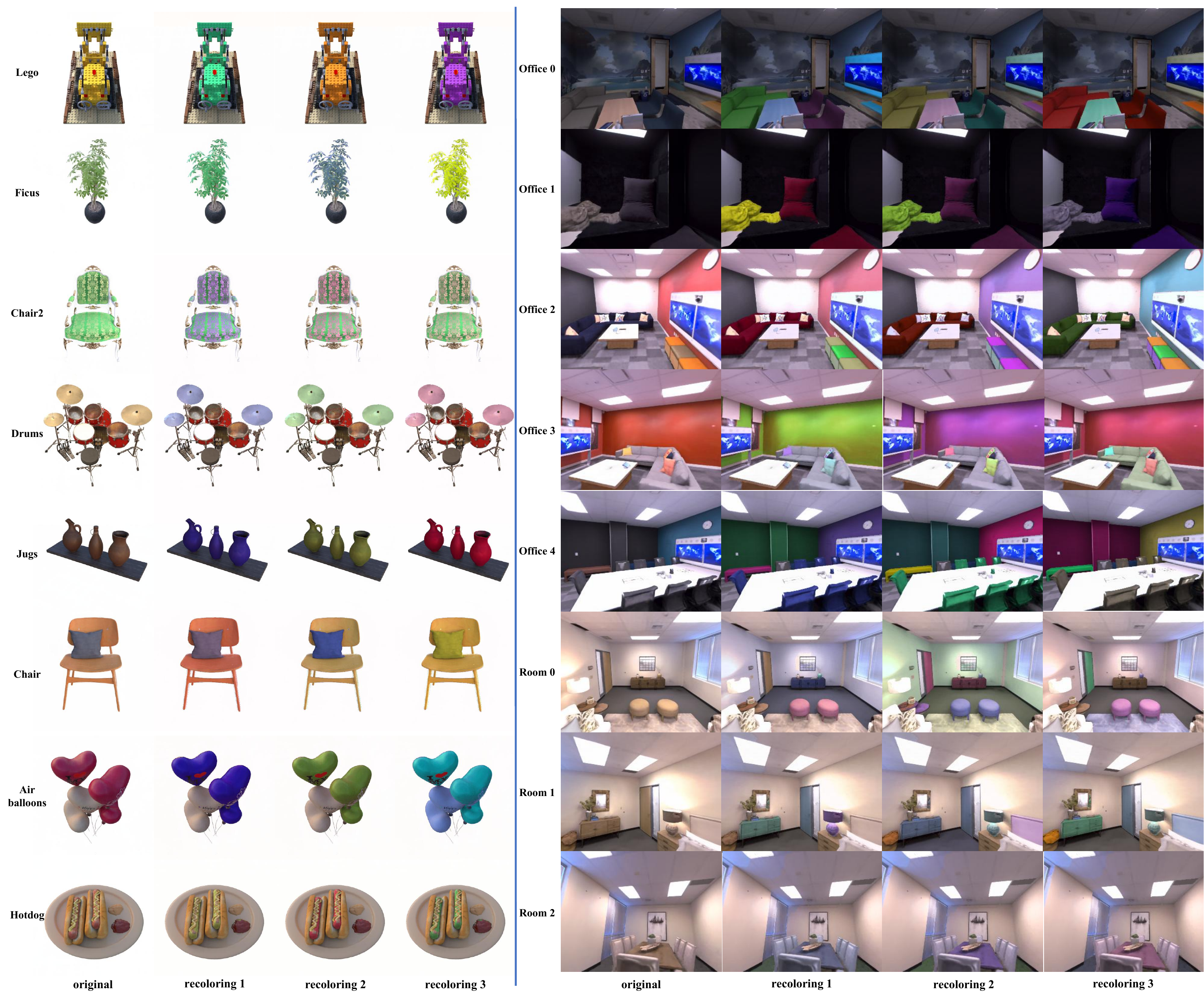}
  \caption{\textbf{Real-Time Scene Recoloring on Synthetic Data.}
  Our approach allows for real-time region-level scene recoloring on synthetic data with a simple user click and selected modified color.}
  \label{supp.fig:Real-Time Scene Recoloring}
\end{figure*}

\begin{figure*}[h]
  \centering
  \includegraphics[width=\linewidth]{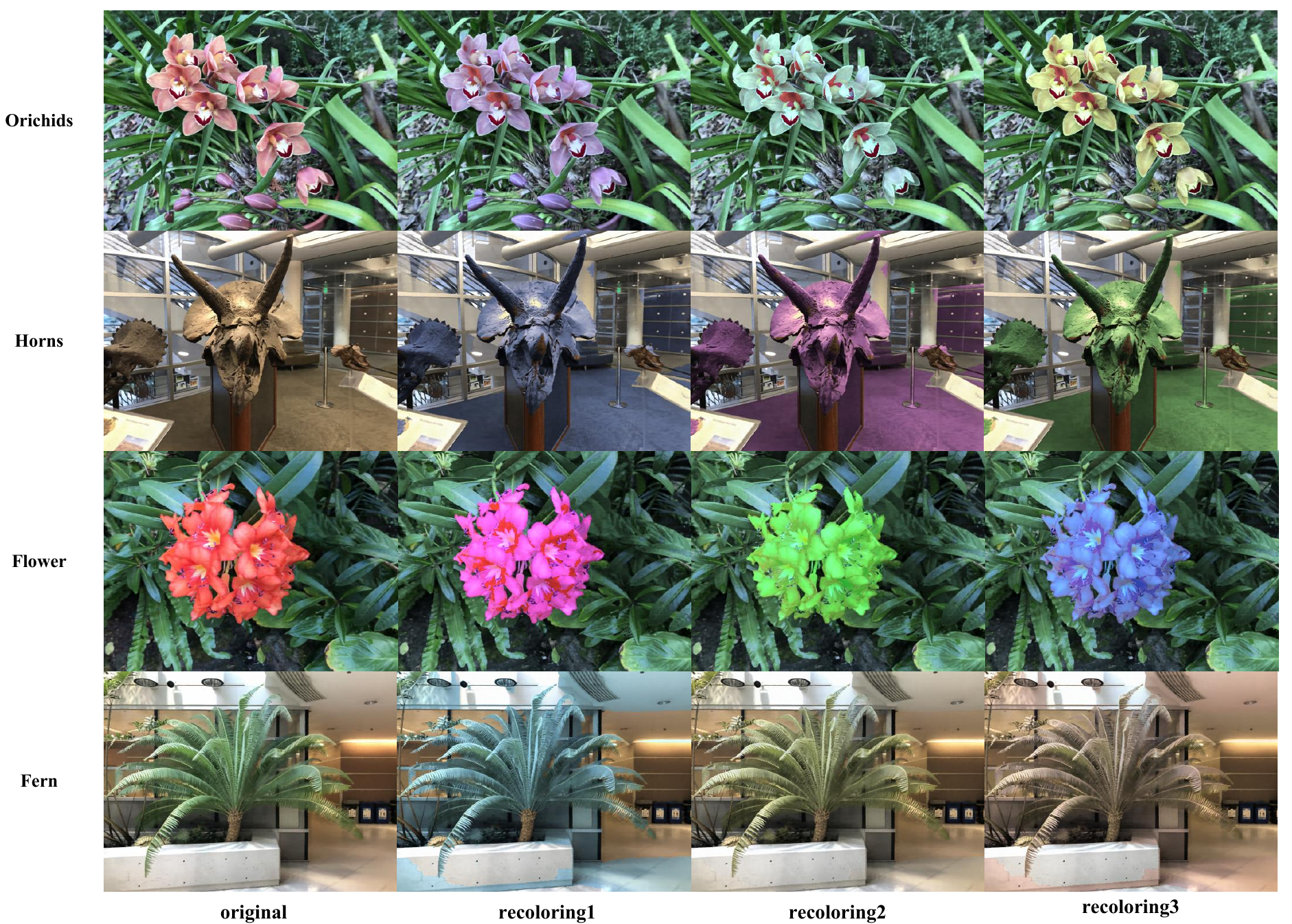}
  \caption{\textbf{Real-Time Scene Recoloring on Real-World Data.}
  Our approach allows for real-time region-level scene recoloring on real-world data with a simple user click and selected modified color.
  }
  \label{supp.fig:Real-Time Scene Recoloring real-world}
\end{figure*}

\noindent\textbf{Real-Time Scene Recoloring.}
\label{Sec:Real-Time Reflectance Recoloring}
The reflectance predicted by the IntrinsicNeRF network is saved as [Semantic category, reflectance category], and the last iteration of the hierarchical iterative clustering method will save the reflectance categories in all semantic categories of the whole scene. Therefore, the [Semantic category, reflectance category] label can be used to quickly find the reflectance value of each pixel point. Based on this representation, we can perform scene recoloring in real-time, just by simply modifying the color of a certain reflectance category, the reflectance values of all pixels in the multi-view images belonging to that category can be modified at the same time, and then the recolored images can be reconstructed using the modified reflectance with the original shading and residual through Eq.~\ref{equa: intrinsic residual}. Fig.~\ref{supp.fig:Real-Time Scene Recoloring} shows the scene recoloring samples on the Blender Object dataset and the Replica Scene dataset. Our method can support semantic recoloring with a simple user click and selected modified color. We also perform scene recoloring on the real-world data to show the generalization ability of our method, shown in Fig.~\ref{supp.fig:Real-Time Scene Recoloring real-world}.


\begin{figure*}[h]
  \centering
\includegraphics[width=\linewidth]{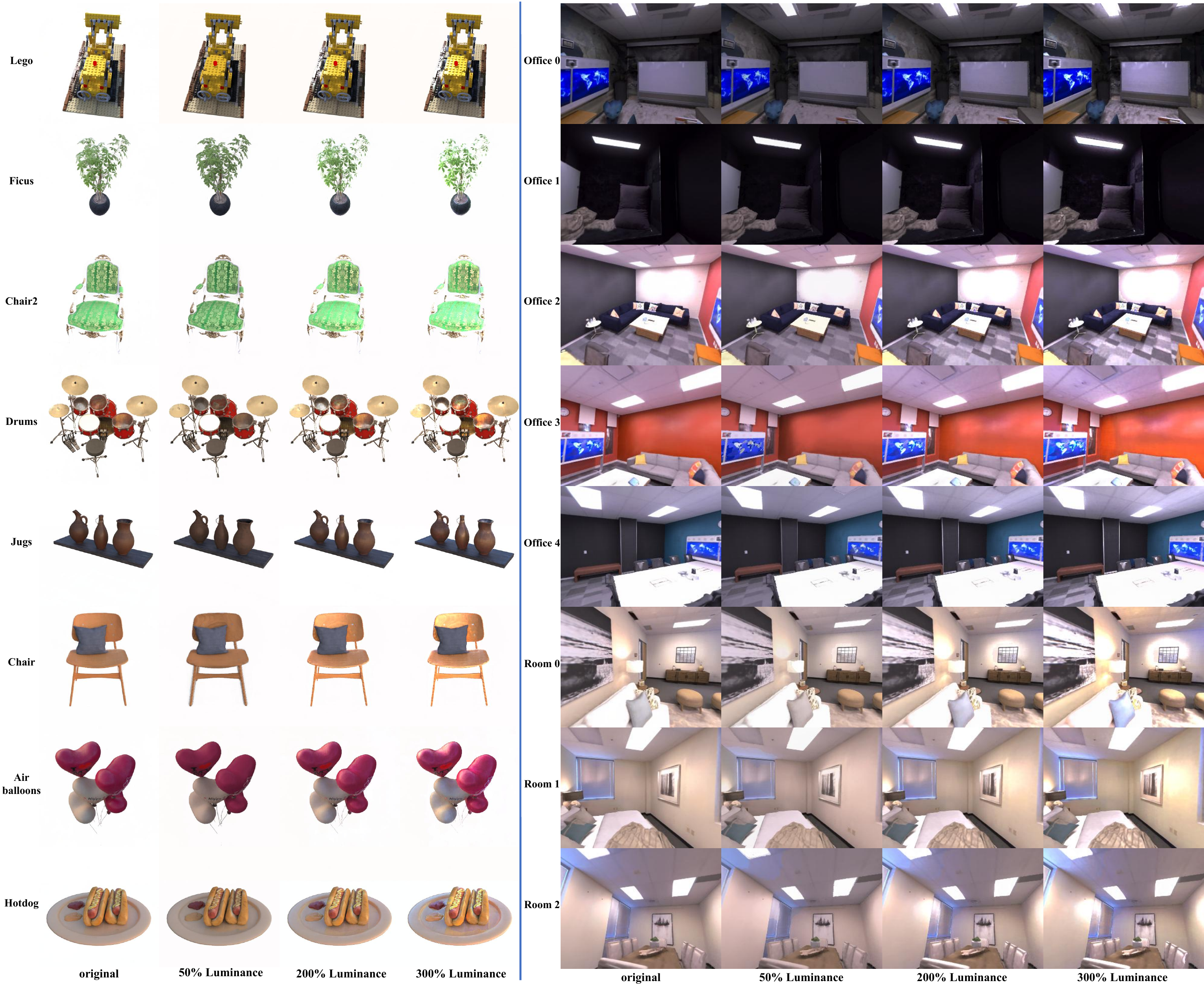}
  \caption{\textbf{Illumination Variation on Synthetic Data.} Left: Blender Object dataset, Right: Replica Scene dataset. We can adjust the brightness of the illumination, which can be applied to the ceiling, sofa, walls, and doors (such as Room 0). Please refer to the supplementary video.}
  \label{supp.fig:illumination variation on Blender Object Dataset}
\end{figure*}

\begin{figure*}[h]
  \centering
\includegraphics[width=\linewidth]{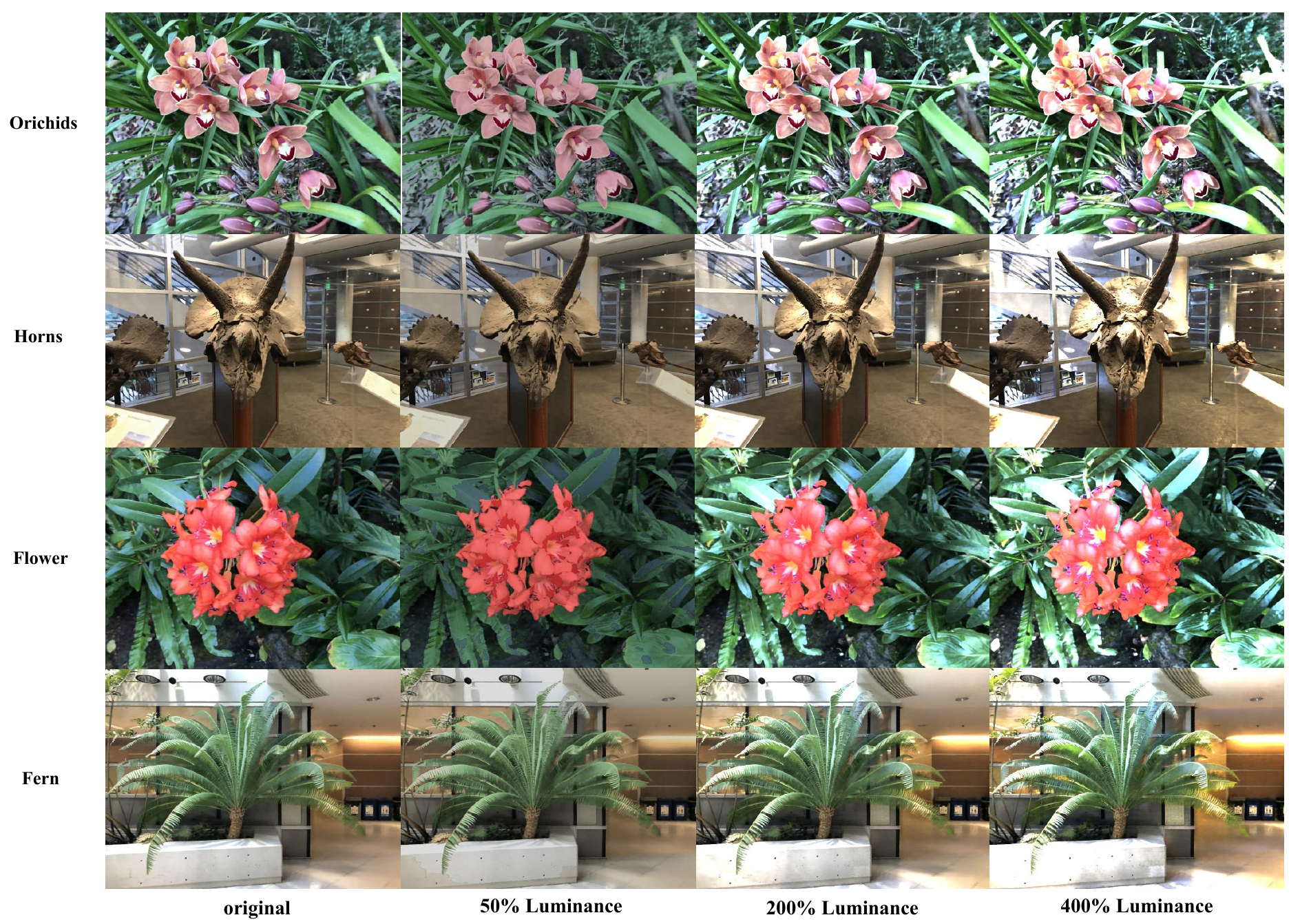}
  \caption{\textbf{Illumination Variation on Real-World Data.} 
  We can adjust the brightness of the illumination on real-world data. Please refer to the supplementary video.
  }
  \label{supp.fig:illumination variation on real-world Dataset}
\end{figure*}

\noindent\textbf{Illumination Variation.}
\label{Sec:Illumination Editing or Relighting}
Since our IntrinsicNeRF can decompose residual terms besides Lambertian assumptions, which may be properties such as specular illumination, we can adjust its overall brightness directly by a multiplicative factor. Specifically, users only need to adjust the sliding buttons of the video editing software and the overall brightness will be modified. We can enhance the light or diminish it, to see the effect of different light intensities, as shown in Fig.~\ref{supp.fig:illumination variation on Blender Object Dataset}. We also perform illumination variation on the real-world data to show the generalization ability of our method, shown in Fig.~\ref{supp.fig:illumination variation on real-world Dataset}.

\begin{figure*}[h]
  \centering
  \includegraphics[width=\linewidth]{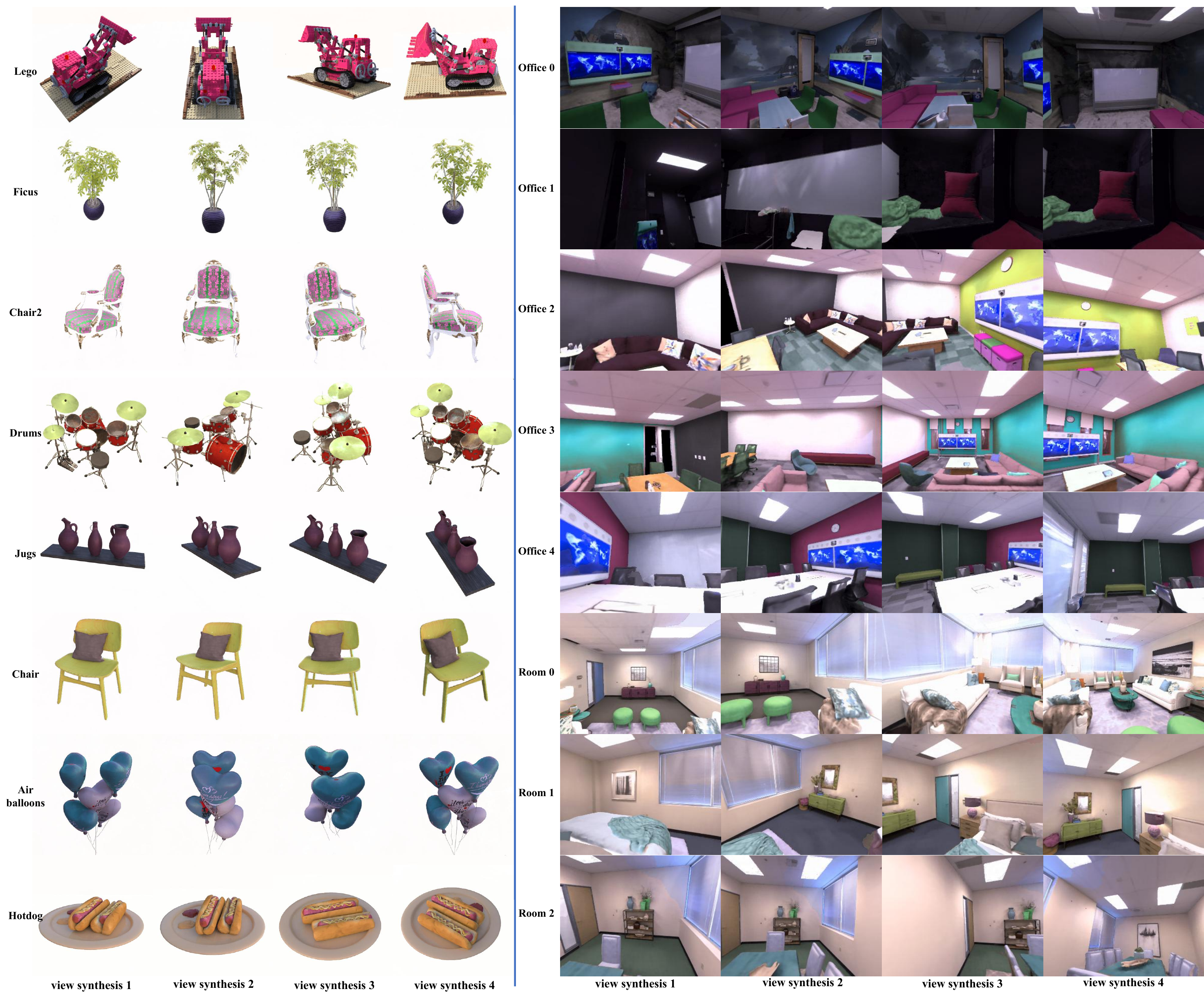}
  \caption{\textbf{Editable Novel View Synthesis on Synthetic Data.} Our method can support real-time augmented editing applications with editable novel view synthesis. Here, we show the view synthesis results with scene recoloring.
  For more details, please refer to the supplementary video.}
  \label{supp.fig: Novel View Synthesis}
\end{figure*}

\begin{figure*}[h]
  \centering
  \includegraphics[width=\linewidth]{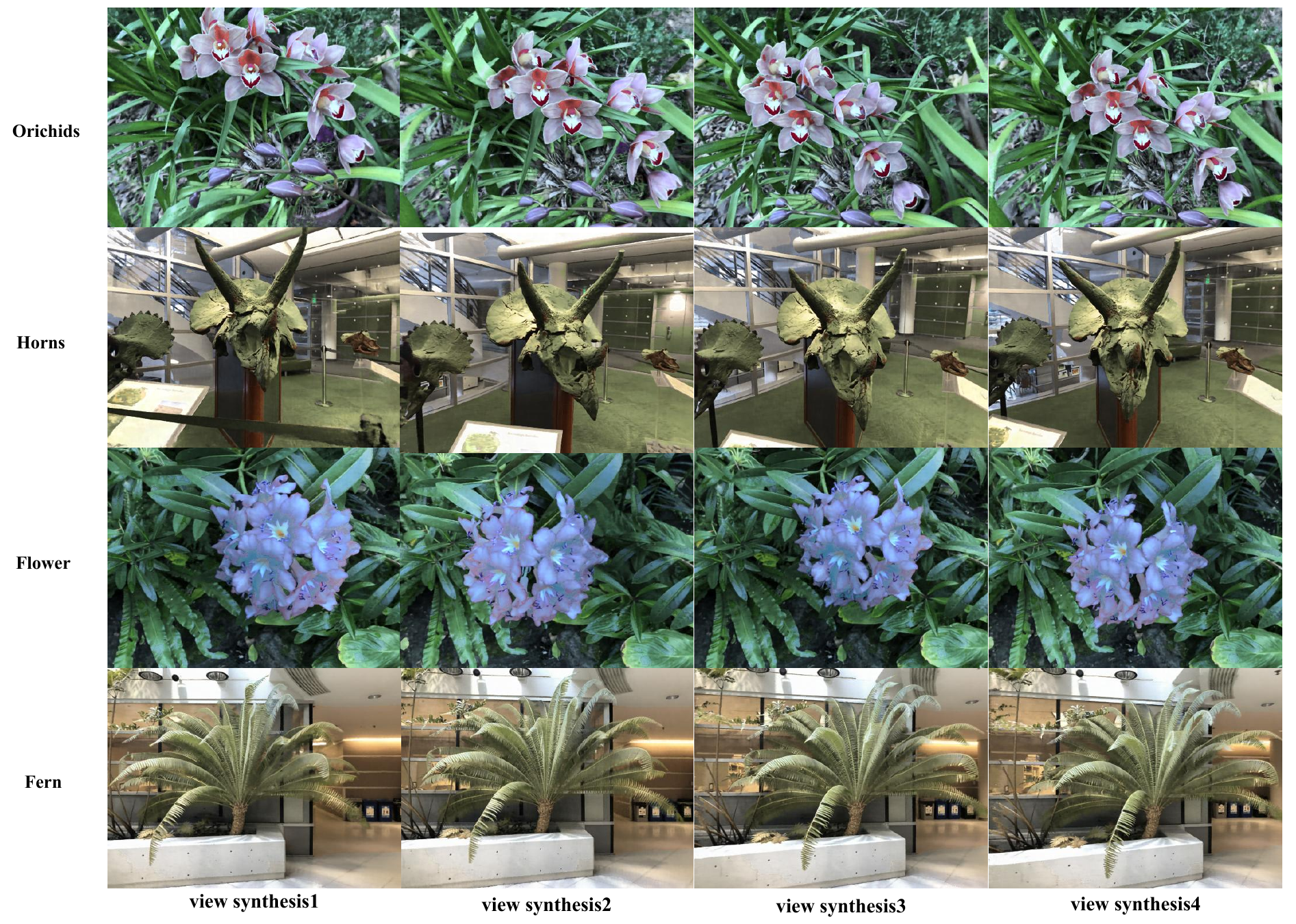}
  \caption{\textbf{Editable Novel View Synthesis on Real-World Data.} 
  Our method can support real-time augmented editing applications with editable novel view synthesis. 
  }
  \label{supp.fig: Novel View Synthesis real-world}
\end{figure*}

\noindent\textbf{Editable Novel View Synthesis.}
\label{Sec:Novel View Synthesis}
Our IntrinsicNeRF gives the NeRF~\cite{mildenhall2020nerf} the ability to model additional fundamental properties of the scene, and the original novel view synthesis functionality is retained. As shown in Fig.~\ref{supp.fig: Novel View Synthesis}, the effects of our video editing application above such as scene recoloring can be applied to the editable novel view synthesis, maintaining consistency. We also perform editable view synthesis on the real-world data to show the generalization ability of our method, shown in Fig.~\ref{supp.fig: Novel View Synthesis real-world}. Please refer to the supplementary video for more details.

\noindent\textbf{Video Editing Software.}
As shown in Fig.~\ref{supp.fig: Video Editing Software}, we visualize the interface of our video editing software, which contains controls for the color palette for the reflectance layer, two sliding bars for shading and residual layers, as well as buttons for playing or recording view synthesis and reset, etc. Due to IntrinsicNeRF with hierarchical clustering and indexing representation, our software can support real-time augmented video editing.

\end{document}